%% file: elsarticle-template-num.tex
\documentclass[preprint,12pt]{elsarticle}

\usepackage{amssymb}

\usepackage{lineno}
\usepackage{xcolor}
\usepackage[]{hyperref}
\usepackage{multirow}
\usepackage{pgfplots}
\pgfplotsset{compat=1.14}
\usepackage{adjustbox}
\usepackage{graphicx}
\usepackage{cellspace}

\journal{Pattern Recognition}

\begin{document}

\begin{frontmatter}



\title{Semi-Automatic Data Annotation guided by Feature Space Projection}


\author[add1]{B\'arbara C. Benato\corref{cor1}}
\ead{barbarabenato@gmail.com}
\author[add1]{Jancarlo F. Gomes}
\ead{jgomes@ic.unicamp.br}
\author[add2]{Alexandru C. Telea}
\ead{a.c.telea@uu.nl}
\author[add1]{Alexandre X. Falc{\~a}o}
\ead{afalcao@ic.unicamp.br}

\cortext[cor1]{Corresponding author}
\address[add1]{Institute of Computing, University of Campinas, Campinas, Brazil}
\address[add2]{Department of Information and Computing Sciences, Faculty of Science, Utrecht University, Utrecht, the Netherlands}

\begin{abstract}
Data annotation using visual inspection (supervision) of each training sample can be laborious. Interactive solutions alleviate this by helping experts propagate labels from a few supervised samples to unlabeled ones based solely on the visual analysis of their feature space projection (with no further sample supervision). We present a semi-automatic data annotation approach based on suitable feature space projection and semi-supervised label estimation. We validate our method on the popular MNIST dataset and on images of human intestinal parasites with and without fecal impurities, a large and diverse dataset that makes classification very hard. We evaluate two approaches for semi-supervised learning from the latent and projection spaces, to choose the one that best reduces user annotation effort and also increases classification accuracy on unseen data. Our results demonstrate the added-value of visual analytics tools that combine complementary abilities of humans and machines for more effective machine learning.
\end{abstract}

\begin{keyword}
Semi-Supervised Learning \sep Unsupervised Feature Learning \sep Interactive Data Annotation \sep Autoencoder-Neural Networks \sep Data Visualization
\end{keyword}

\end{frontmatter}


\section{Introduction}
\label{s.intro}

Machine Learning (ML) models have been extensively investigated and used for regression and classification problems~\cite{Krizhevsky:2012,Kim_2016_CVPR,Levine:2016}. More recently, Convolutional Neural Networks (CNNs) have shown great success in many applications, such as image/text classification~\cite{Lecun:98} and speech recognition~\cite{Hinton:2012}, since they require considerably less effort to optimize parameters than the common feature extraction pipeline\,\cite{Lecun:98}. However, CNNs may require a high number of labeled samples (annotated objects) for training\,\cite{Yosinski:2014}. 

While small labeled training sets can impair the ability of an ML model to correctly classify new samples (a problem known as \emph{over-fitting}\,\cite{Srivastava:14}), large unlabeled sets make visual inspection and annotation very expensive for the expert. Human costs become even so more prohibitive in domains that require specialized knowledge about the objects, like Medicine and Biology. Solutions for small labeled sets include data augmentation\,\cite{Mash:2016} and regularization methods\,\cite{Nowlan:1992}. For large unlabeled sets, semi-supervised classifiers have been used to propagate labels from a small supervised set to the many unsupervised samples by exploring the sample distribution in some feature space\,\cite{Kingma:2014, forestier:2016, Papernot:2017}. Yet, none of these approaches has combined the cognitive ability of humans in data abstraction with the ability of machines in data processing to increase the number of labeled objects. 

Recent studies have investigated the use of feature space projections and visual analytics to understand and engineer ML models\,\cite{RauberInfVis2017, RauberVGTC2016,Peixinho:2018,Bernard:2018,BenatoSibgrapi:2018}. Such work addresses both aforementioned labeling cases with approaches for interactive data augmentation~\cite{Peixinho:2018} and interactive data annotation~\cite{Bernard:2018,BenatoSibgrapi:2018} guided by feature space projections, respectively. Bernard et al.\,\cite{Bernard:2018} have compared interactive data annotation in a feature space projection with an active learning technique, in which experts supervise and annotate samples selected by a classifier and the classifier is retrained to annotate and select more samples in the original feature space. They discovered that interactive data annotation in the feature space projection is superior to active learning. Benato et. al.~\cite{BenatoSibgrapi:2018} have showed that when the user propagates labels to a large unsupervised sample-set guided by the true-label knowledge of a few samples and by the visual information of the sample distribution in a feature space projection, the resulting labeled training-set is more correct than the one created by semi-supervised classifiers in the original feature space. Hence, classifiers trained from such interactively labeled sets can better predict labels of unseen test samples than those trained from automatically labeled sets. Yet, Bernard et al.~\cite{Bernard:2018} and Benato et al.~\cite{BenatoSibgrapi:2018} have not \emph{combined} automatic and interactive approaches for label propagation --- i.e, they have not been concerned with the user \emph{effort} in visual data inspection and annotation.

In this work, we fill the above gap by proposing a semi-automatic approach that reduces user labeling effort while achieving better classification accuracy on unseen test sets. For this, we exploit the concept of \emph{sample informativeness} from Active Learning (AL). Such approaches select samples for expert supervision based on their informativeness --- i.e., potential to improve the design of a classifier from the knowledge of their true label\,\cite{Settles:2009}, measured by the \emph{confidence} of a classifier about the label assigned to a sample\,\cite{Patra:2012,Miranda:2009,Spina:2012,Tavares:2012}. In our case, we propagate labels to samples with high-confidence values; and enable the expert focus on low-confidence values for manual label propagation. For this, the user visually analyzes the sample distribution in a 2D scatterplot created by the \emph{t-Distributed Stochastic Neighbor Embedding} (t-SNE) technique\,\cite{MaatenJMLR:14}, constructed similarly to\,\cite{Bernard:2018,BenatoSibgrapi:2018}, and the true-label knowledge of only a few samples per class. Although our method can explore further classifier improvement of the classifier by multiple iterations of AL with additional supervised samples, we solve data annotation from a single user interaction for label propagation with no sample supervision. For automatic label propagation, we evaluate two semi-supervised classifiers trained in both latent and projection spaces for automatic label estimation and choose the best one for our goal. We show that our semi-automatic label propagation (SALP) method achieves end-to-end better classification results as compared to both fully automatic label propagation and fully manual label propagation.

This work is organized as follows. Section~\ref{s.method} presents our semi-supervised data annotation approach. Section~\ref{s.experiments} presents the experimental setup, compared baselines, used datasets, and experimental results. Section~\ref{s.discussion} discusses our results. Section~\ref{s.conclusion} concludes the paper.

\section{Semi-Automatic Projection-Based Data Annotation}
\label{s.method}
\begin{figure}[h]
\label{f.pipeline}
\centering
\includegraphics[width=1.0\linewidth]{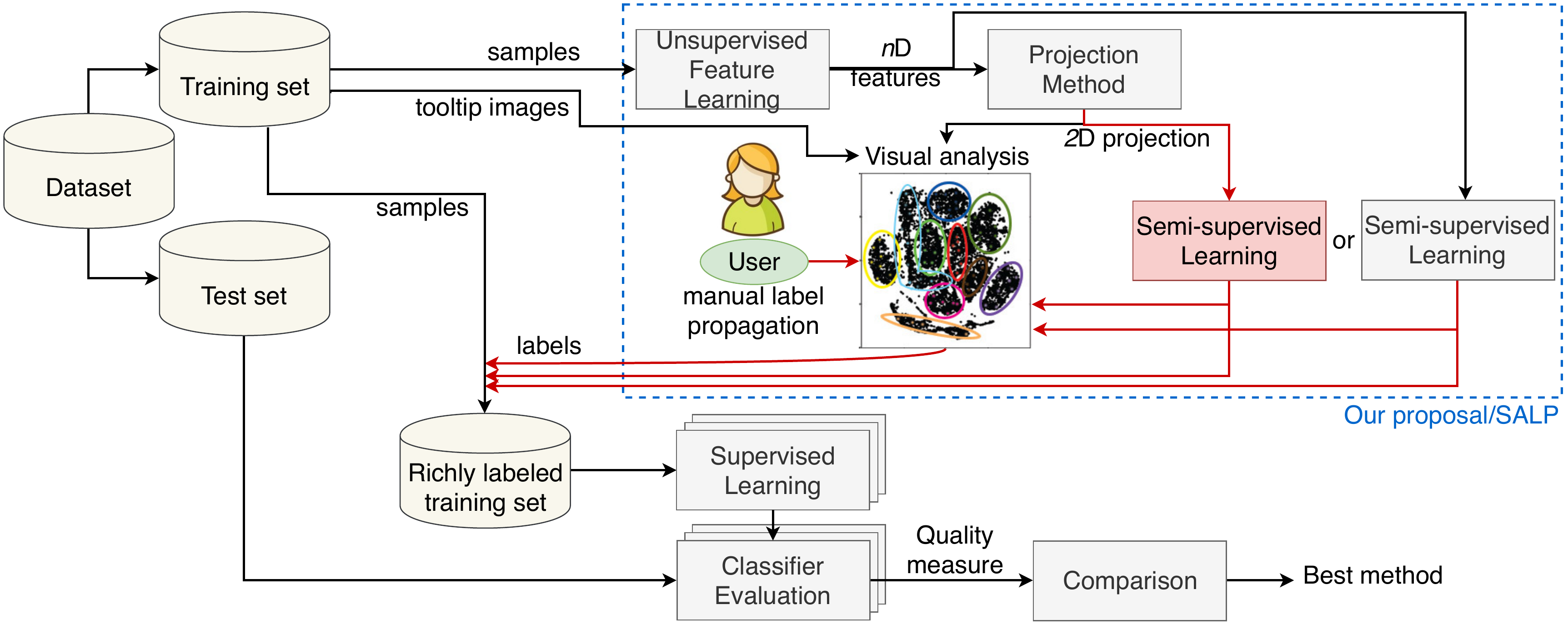}
\caption{Semi-automatic data annotation pipeline. We extract features by unsupervised learning from the training set and next use these to project this set to a 2D scatterplot. We next enrich the training set by propagating labels from supervised to unsupervised samples by automatic methods (in both latent and projection spaces) and by manual user-controlled methods. We finally compare the quality of the classifiers trained on such training sets to decide on the best label propagation method. Red indicates additions to earlier related work\,\cite{BenatoSibgrapi:2018}.
}
\end{figure}
Given a training set with a low number of supervised samples and a considerably larger number of unsupervised samples, our semi-automatic data annotation approach has four steps: 

\begin{itemize}

\item \emph{unsupervised feature learning:} We start by extracting features from the input dataset. To minimize the number of supervised samples needed, we adopt an unsupervised feature-learning procedure  (Sec.~\ref{s.unsupervised});

\item \emph{feature space projection:} We create a feature space 2D projection that captures well the sample distribution in the latent feature space for further visual analysis;

\item \emph{semi-supervised label estimation:} We propagate labels automatically to high-confidence unlabeled samples, thereby increasing with training-set size with little effort and high quality (Sec.~\ref{s.semi_sup});

\item \emph{visual analysis:} The expert creates additional labeled samples to the above ones, by interactively propagating labels to the less-confident samples using the 2D projection (Sec.~\ref{s.ilp}).
\end{itemize}

\subsection{Unsupervised Feature Learning}
\label{s.unsupervised}
We use an Autoencoder Neural Network (AuNN)\,\cite{Masci2011, Vincent} for unsupervised feature learning. AuNNs consist of two parts, encoder and decoder. The encoder maps the input samples to points in a reduced (latent) feature space; the decoder reconstructs these samples. The two parts are coupled and trained together by backpropagation. As cost function, we use the mean squared error between the original and reconstructed samples. For small errors, the obtained latent feature space is a reasonable representation of the original sample distribution. Hence, we train the AuNN with all labeled and unlabeled samples by ignoring labels. After evaluating several models, we decided for a Stacked Convolutional AuNN\,\cite{Masci2011} --- a neural network that presents convolutional layers and can usually obtain relevant latent features. For our experiments, we use image datasets. However, this latent feature learning can be used for any other kind of data that can be suitably mapped to the input layer of the encoder. Section~\ref{s.experiments} presents implementation details.

\subsection{Feature Space Projection}
\label{s.featspace}
Previous works indicate that 2D projections, created by the t-SNE algorithm\,\cite{Hinton:2006,Maaten:2008}, achieve this goal well\,\cite{RauberInfVis2017,Bernard:2018,BenatoSibgrapi:2018}, so we follow these (Sec.~\ref{s.featspace}).

The dimension of the latent feature space can still be considered very high (with usually hundreds to thousands of features) and so unfeasible for visual inspection of the sample distribution. As previously mentioned, we wish to reduce the latent space to two dimensions by preserving as much as possible the relevant structure of the data. The most suitable techniques for this task seem to preserve local distances between samples and the t-SNE algorithm satisfies this criterion~\cite{MaatenJMLR:14}. It is a non-linear projection that depends on the choice of two parameters: perplexity and number of iterations. Our choice for these parameters is discussed in Section~\ref{s.experiments}.  

\subsection{Semi-Supervised Label Estimation}
\label{s.semi_sup}
For semi-supervised label estimation, we consider two techniques that explore the sample distribution in a given feature space to propagate labels with confidence values from supervised to unsupervised ones: Laplacian Support Vector Machines (LapSVM)\,\cite{Sindhwani:2005,Belkin:2006} and Semi-Supervised Classification by Optimum-Path Forest (OPF-Semi)\,\cite{Amorim:2016}. We evaluate both methods on both latent and projection spaces. Given that the performance of OPF-Semi in label propagation is much higher than that of LapSVM (see Sec.~\ref{s.experiments}), we select OPF-Semi to output confidence values, used next for our manual label propagation (Sec ~\ref{s.ilp}). Additionally, we found that OPF-Semi in the projection space outperforms itself in the latent feature space (see Sec.~\ref{s.experiments}). Hence, we use the 2D version of OPF-Semi for semi-automatic data annotation.

\begin{figure}[htb]
\centering
  \includegraphics[width=.45\linewidth]{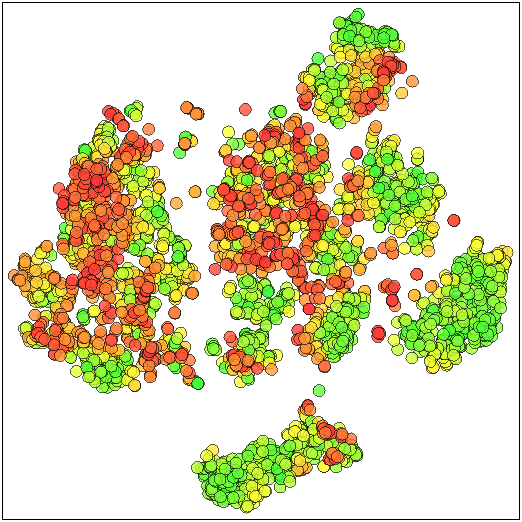}\\
  \caption{Feature projection showing unsupervised samples from red (low confidence) to green (high confidence).}
  \label{f.colormap}
\end{figure}

OPF-Semi maps (un)labeled samples to nodes of a graph and computes an optimum-path forest rooted at labeled samples. In this forest, each node $s$ is conquered (labeled) by the root $R$ that offers a path of minimum cost $k(R,s)$ to $s$. We use costs to compute label confidence values $c(s)$ as described in\,\cite{Miranda:2009,Spina:2012,Tavares:2012}. In brief: Let $A$ and $B$ be two roots for sample $s$ so that $A$ is the one that has conquered $s$ ($k(R,s)$ is minimal) and $B$, having a different label than $A$, offers the second-best cost $k(B,s)$ to $s$. We assign the confidence $c(s) = 1-k(A,s)/ (k(A,s)+k(B,s))$, $c(s) \in [0,1]$, to the label of $s$ given by $A$. That is, if the second-best cost $k(B,s)$ is much larger than the minimal cost $k(A,s)$, the label $A$ has a high confidence. We use the confidence as follows: All labels assigned by OPF-Semi having a confidence above a threshold $\tau$ are used as such in the training process. The threshold $\tau$ is chosen by the user based on the visual analysis of the feature projection with unsupervised samples colored by their confidence values from red (low $c$) to green (high $c$) (Fig.~\ref{f.colormap}). Changing $\tau$ interactively by a slider lets the user (a) say that high-confidence samples can keep their likely good labels assigned by OPF-Semi and (b) focus on the remaining low-confidence samples to assign them labels by manual label propagation, described next in Sec.~\ref{s.ilp}. Users can choose the exact threshold $\tau$ balancing how much they wish to trust OPF-Semi \emph{vs} how many samples they are willing to label manually.

\subsection{Manual Label Propagation}
\label{s.ilp}
\begin{figure}[htb]
\centering
  \includegraphics[width=.45\linewidth]{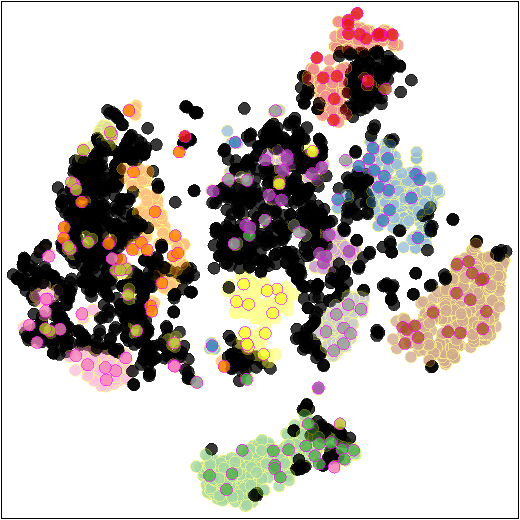}\\
  \caption{Semi-automatic label propagation is done from the supervised samples (points colored by class, saturated colors, red border) first automatically to the unsupervised and high-confidence ones (light colors, no border). Remaining low-confident samples (black) are candidates for manual propagation.}
  \label{f.proj}
\end{figure}

The added value of user-driven label propagation in a t-SNE projection was demonstrated by the interactive label propagation technique in\,\cite{BenatoSibgrapi:2018} which we refer to next as ILP for brevity.
However, ILP propagation is fundamentally affected by the quality of the latent features extracted by the AuNN (Sec.~\ref{s.unsupervised}) \emph{and} the quality of the t-SNE projection itself: If both these operations faithfully preserve the similarity of original samples, then the user can likely propagate labels well, by simply selecting points close in the projection to the supervised samples. If either the latent space or the projection create errors, which they inherently do\,\cite{nonato18}, this will likely create wrong labels. We assist the user in this process as follows. We color the supervised points in the projection by their labels, and color all low-confidence unsupervised points $s$ having $c(s) < \tau$ in black (Fig.~\ref{f.proj}). The black points are projected before the colored points, in order to minimize undesired occlusions. When moving the mouse pointer over a projected point, we show its sample image in a tooltip. The user next employs these three sources of information -- proximity of unsupervised (black) points in the 2D projection to supervised (colored) ones, low-confidence value of the unsupervised points, and similarity of unsupervised-to-supervised tooltip images -- to decide which unsupervised samples get which supervised label. Label propagation is next done simply by selecting desired points in the projection and clicking to assign them a supervised-point label. 

\section{Experiments and Results}
\label{s.experiments}
We next present the experimental setup, baselines, datasets, implementation details, and  experimental results used for validating our semi-automatic data annotation method.

\subsection{Experimental Setup}
\label{s.expsetup}
We divide each available dataset $D$ into three subsets for validation:  a very small training set $S$ with a few supervised samples per class ($3\% |D|$); a considerably larger training set $U$ with unsupervised samples for label propagation ($67\% |D|$); and a set $T$ with unseen test samples ($30\% |D|$). Next, based on the user-chosen confidence threshold $\tau$, we split $U$ into high-confidence samples $L_c$, which get their label from OPF-Semi, and low-confidence ones $L_i$, which can be interactively labeled by the user. Note that $L_c \cap L_i = \emptyset$ and $L_c \cup L_i \neq U$, since the user can choose not to label $L_i$ entirely, to minimize manual labeling effort. We randomly split $D$ into $S$, $U$, and $T$ this way three times and repeat the evaluation --- i.e., label propagation from $S$ to $U$ followed by supervised training on $S\cup U$ and testing on $T$ --- for statistical purposes.

After labels are propagated from $S$ to $U$, we train a supervised classifier on $S\cup U$ using the latent feature space. For this task, we used the Optimum-Path Forest (OPF)\,\cite{Papa:2012} and Support Vector Machines (SVM)\,\cite{Hearst:1998}. OPF has no hyperparameters to set, so it is simple to use. For SVM, we find optimal values for its hyperparameters $\sigma$ (influence radius), $C$ (regularization) and kernel type by grid search over the ranges 
$[0.1, 0.000001]$, $[1, 10000]$ and the kernel functions \textit{Gaussian radial basis} and \textit{linear} respectively, using $3$ splits and stratified random sampling with $70\%$ and $30\%$ of the samples from $S\cup U$ used for training and validation, respectively. We test the classifiers on $T$ and measure their effectiveness by Cohen's $\kappa$ coefficient\,\cite{Cohen:1973}. The $\kappa$ coefficient is within $[-1,1]$, where $\kappa \leq 0$ means no agreement and $\kappa=1$ means complete agreement between two annotators. Additionally, we also compute the accuracy of label propagation on $U$ for each approach, that is the number of labeled samples correctly assigned divided by the number of unsupervised samples ($|U|$). Therefore, the best approach for label propagation is the one that produces the best supervised classifiers. Since we consider the $\kappa$ as effectiveness measure, the best supervised classifier is then the one that provides the best $\kappa$ result.

\subsection{Baselines}
\label{ss.baselines}
As described in Section~\ref{s.method}, we propose a semi-automatic label propagation (SALP) that uses OPF-Semi in the 2D t-SNE projection space to propagate labels to high-confidence samples and the user to propagate labels to low-confidence samples, respectively. We next compare SALP with the following three baselines:
\begin{enumerate}
\item \emph{No label propagation (NLP):} SVM and OPF, are trained from only $S$, ignoring set $U$.
\item \emph{Automatic label propagation (ALP):} set $U$ is fully labeled by one of the four ALP methods below and SVM and OPF are trained from $S\cup U$. 
\begin{enumerate}    
    \item LapSVM using the $n$D latent feature space.
    \item LapSVM using the 2D t-SNE projection space.
    \item OPF-Semi using the $n$D latent feature space.
    \item OPF-Semi using the 2D t-SNE projection space.
\end{enumerate}
\item \emph{Interactive label propagation (ILP):} set $U$ is fully labeled by the user and SVM and OPF are trained from $S\cup U$, as in\,\cite{BenatoSibgrapi:2018}.
\end{enumerate}

In all above cases, we test SVM and OPF on $T$.

\subsection{Datasets}
\label{ss.datasets}
\begin{figure}[htb]
\centering
  \includegraphics[width=0.7\linewidth]{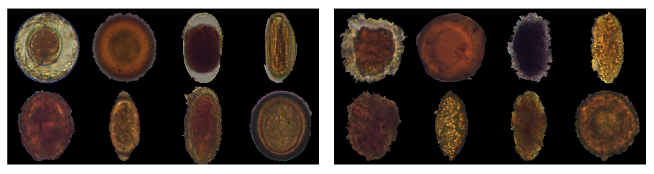}\\
  \caption{Examples of each species of H.Eggs (left) and similar images of impurities (right).}
  \label{f.parasites}
\end{figure}

Our first dataset contains 5000 images ($28\times28$ pixels each) of handwritten digits from 0 to 9, randomly selected from the popular public dataset MNIST\,\cite{Lecun:2010:mnist}. Our next three datasets use images ($200 \times 200$ pixels each) from an automatic processing pipeline that separates microscopy images of human intestinal parasites into three groups: (i) \emph{Helminth larvae} and fecal impurities ($3514$ images); (ii) \emph{Helminth eggs} and fecal impurities ($5112$ images); and (iii) \emph{Protozoan cysts} and fecal impurities ($9568$ images).Fecal impurity is a diverse class that has very similar samples to parasites (see Fig.~\ref{f.parasites}). We consider these three datasets with and without images of fecal impurities, yielding five datasets for testing our proposal, apart from MNIST. The number of classes in each dataset is as follows: (i) H.Larvae has two categories; (ii) H.Eggs has nine categories (\emph{H.nana}, \emph{H.diminuta}, \emph{Ancilostomideo}, \emph{E.vermicularis}, \emph{A.lumbricoides}, \emph{T.trichiura},  \emph{S.mansoni}, \emph{Taenia}, and impurities); and (iii) P.cysts has seven categories (\emph{E.coli}, \emph{E.histolytica}, \emph{E.nana}, \emph{Giardia}, \emph{I.butschlii}, \emph{B.hominis}, and impurities), respectively. Those are the most common species of human intestinal parasites in Brazil, which are responsible for public health problems in most tropical countries~\cite{Suzuki:2013}. All three datasets are unbalanced with considerably more impurity samples. The images of parasites have been annotated by biomedical specialists. Table~\ref{t.split_data} gives the number of images in each set $S$, $U$, and $T$ after the random split described in Sec.~\ref{s.expsetup}.

\input{split_data}

\subsection{Implementation Details}
\begin{figure}[htb]
\centering
  \includegraphics[width=0.68\linewidth]{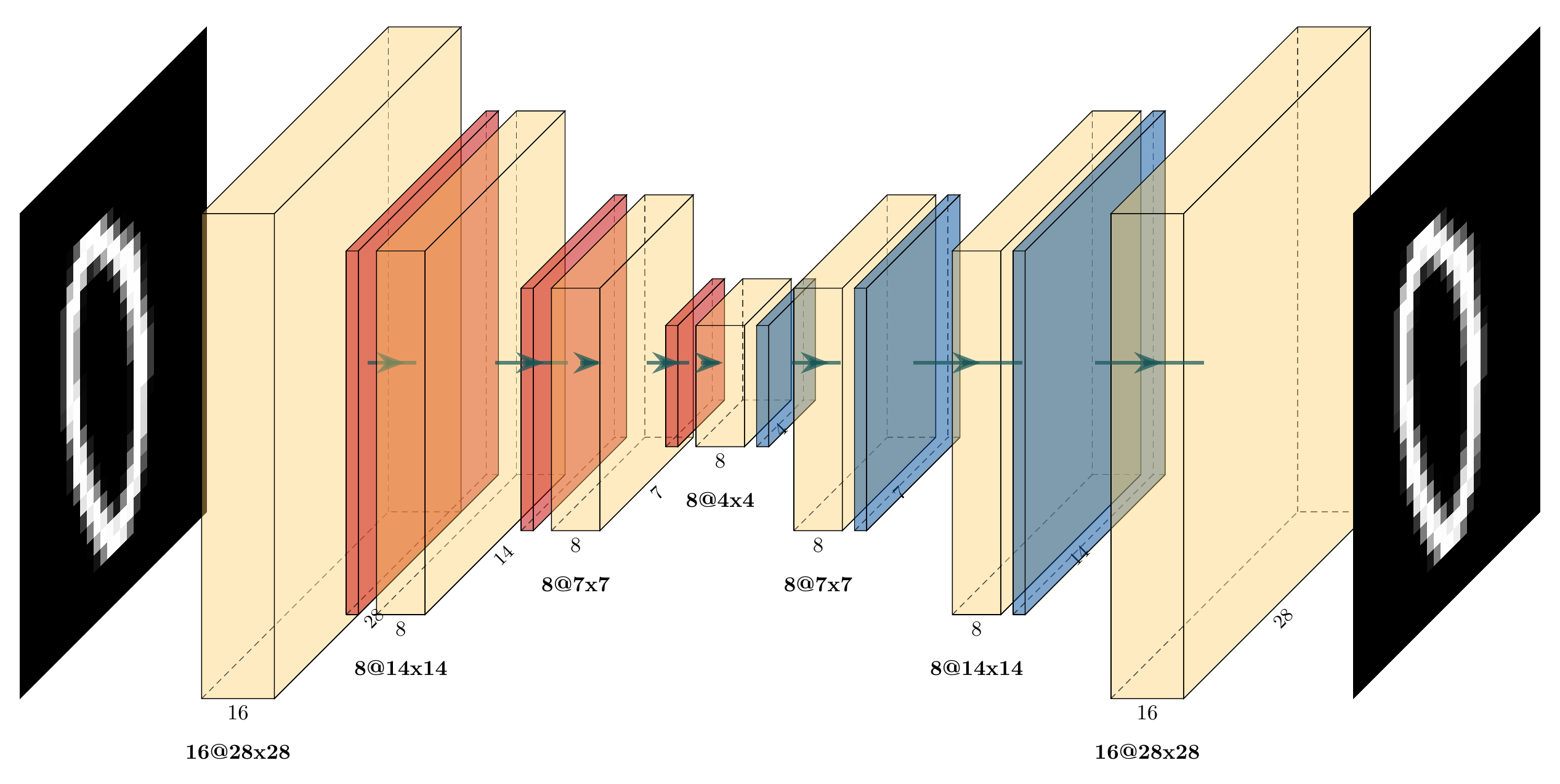}
  \hspace{0.10cm}
  \includegraphics[width=0.7\linewidth]{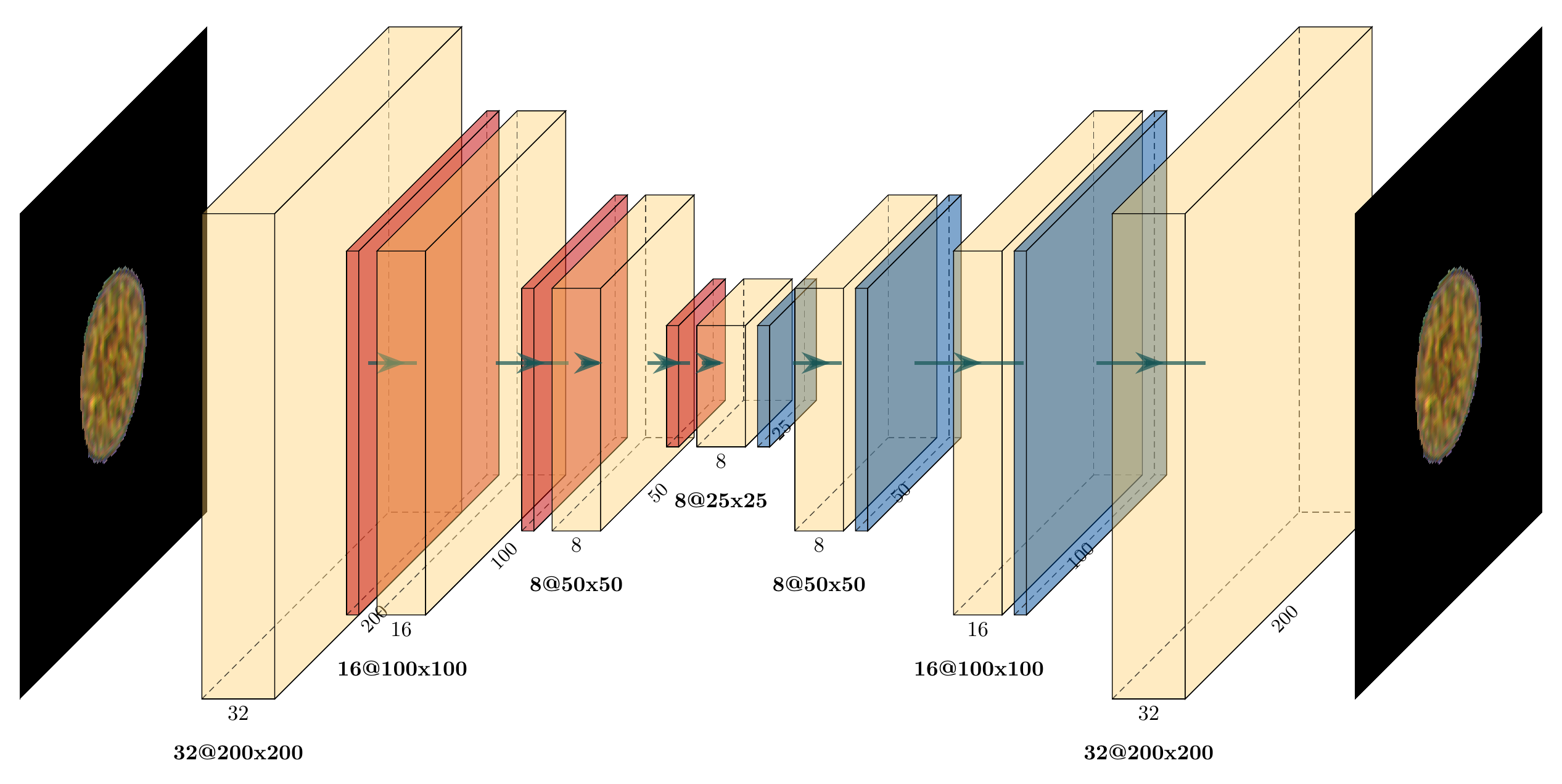}\\
  \caption{AuNN architecture for MNIST dataset (top) and for Parasites datasets (bottom). The yellow layers are the convolutional layers, the red layers at the beginning of each network are the Max Pooling layers and the blue layers are the Up-Sampling layers.}
  \label{f.arch}
\end{figure}

\noindent\textbf{Feature extraction:}  Figure~\ref{f.arch} shows the AuNN architectures for the MNIST and parasites datasets. We implemented these networks in Keras\,\cite{Chollet:2015:keras} with $6$ convolutional layers of $3\times 3$ filters,  $3$ for the encoder and $3$ for the decoder, respectively. After each convolutional layer, we use \textit{ReLU} activation and apply max-pooling in the encoder and upsampling in the decoder. We normalize the input images within $[0,1]$, since the output requires sigmoid activation. We choose the number of filters based on the dataset: For MNIST, the 6 convolutional layers use $16$, $8$, $8$, $8$, $8$, and $16$ filters. For the 5 parasites datasets, we use $32$, $16$, $8$, $8$, $16$,  and $32$ filters respectively. As cost function, we use mean squared error as it provides more suitable results in reconstruction task with fewer training epochs. We use 50 epochs for the easier datasets (MNIST and H. Eggs without impurities) and 100 for the others. For MNIST, we use a latent feature space of $n=128$ dimensions. For the parasites, which have higher-resolution and more complex images, we use $n=5000$ dimensions.

\noindent\textbf{Projection:} Different choices of t-SNE parameters can lead to different 2D projections\,\cite{wattenberg16}. We found empirically that for a range of  $1000$ to $7000$ samples in $S\cup U$, setting t-SNE's perplexity to $40$ and maximum iteration count to $1000$ respectively yields good projections for label propagation.  

\subsection{Experimental Results}
We discuss the performance of our pipeline, measured by the performance of the classifiers trained from $S\cup U$ in the latent feature space and tested on $T$, by answering the following questions:
\begin{itemize}
\item Which space ($n$D latent, 2D projection) is better for ALP? (Sec.~\ref{sec:q1})
\item How to set the confidence threshold $\tau$? (Sec.~\ref{sec:q2})
\item Which approach (manual, semi-automatic, automatic) best propagates labels from $S$ to $U$? (Sec.~\ref{sec:q3})
\item What is the end-to-end value of SALP? (Sec.~\ref{sec:q4})
\item How do results depend on the projection quality? (Sec.~\ref{sec:q5})
\end{itemize}

Note that we use the 2D projection space only for manual label propagation, i.e. not for testing, since we cannot assume that set $T$ is known during training. 

\subsubsection{Influence of reducing the feature space from $n$D to 2D}
\label{sec:q1}
Table~\ref{t.results1} presents mean and standard deviation values of Cohen's $\kappa$ for classifiers on set $T$ for each dataset, as well as the sizes of $S$, $U$, and $S\cup U$, and the mean accuracy values in automatic label propagation for LapSVM and OPF-Semi, used in the $n$D feature space and also in the 2D projection space, as well as the option of not propagating labels. We get several insights. First, we see that LapSVM performs sometimes better and sometimes worse in $n$D as compared to 2D, depending on the dataset. In contrast, OPF-Semi consistently shows a positive impact of reducing the feature space independently of the dataset. This happens even when its label-propagation performance is not the best one. 
\input{results_1_new}

\subsubsection{The choice of the confidence threshold}
\label{sec:q2}
\begin{figure}[htbp!]
\newcommand\cincludegraphics[2][]{\raisebox{-0.4\height}{\includegraphics[#1]{#2}}}
\centering
\vspace{-1.0cm}
\setlength{\tabcolsep}{1.5pt}
\begin{tabular}{cc|ccc}
  & & \footnotesize $S\cup U$ & \footnotesize $L_c$ & \footnotesize $U\setminus L_c$\\
  \hline
  \rotatebox[origin=c]{90}{{\footnotesize MNIST}} &
  \rotatebox[origin=c]{90}{\footnotesize $\tau=0.5$} &
  \cincludegraphics[width=.2\linewidth]{figs/proj/mnist_3.png} & \cincludegraphics[width=.2\linewidth]{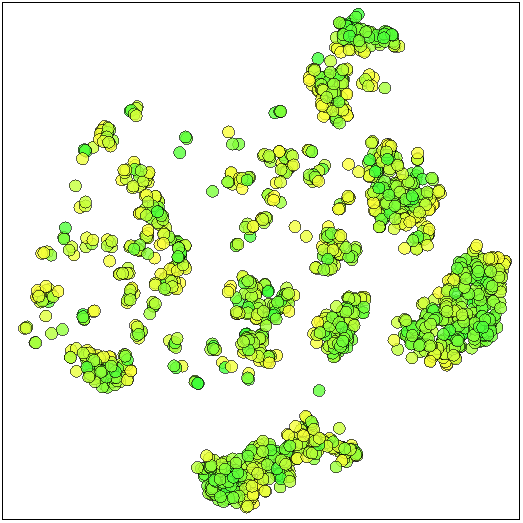} & \cincludegraphics[width=.2\linewidth]{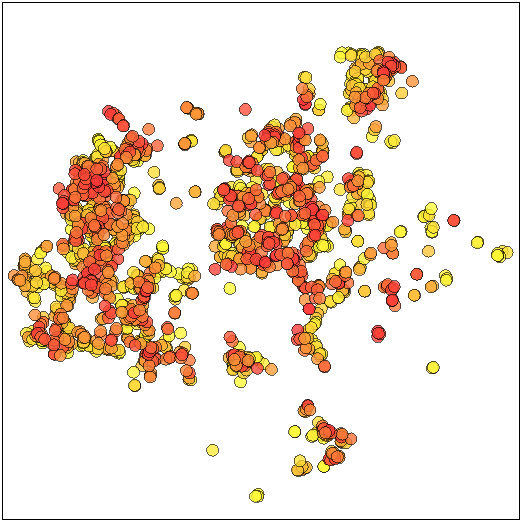}\\
  \rotatebox[origin=c]{90}{\footnotesize H.Eggs} &
  \rotatebox[origin=c]{90}{\shortstack[l]{\footnotesize $\tau=0.6$}} &
  \cincludegraphics[width=.2\linewidth]{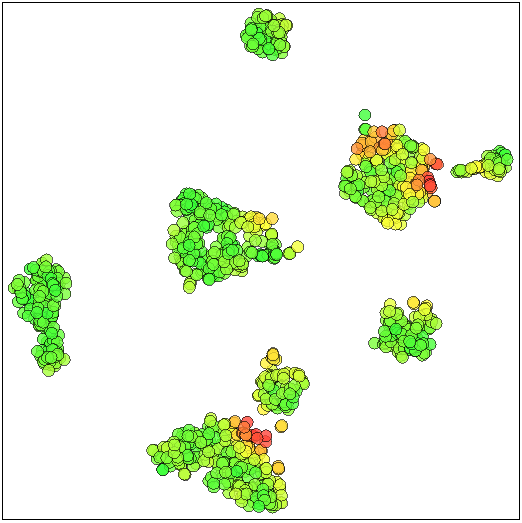} & \cincludegraphics[width=.2\linewidth]{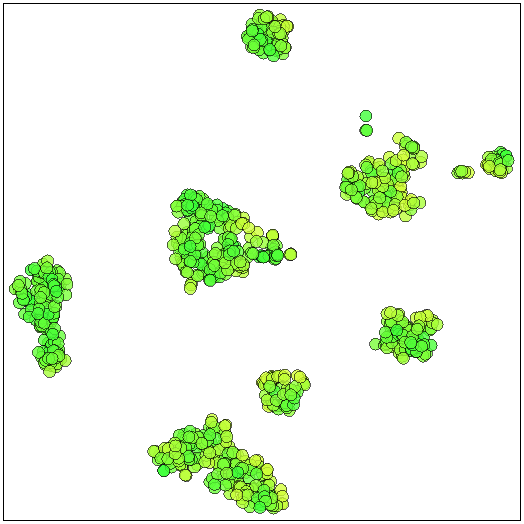} & \cincludegraphics[width=.2\linewidth]{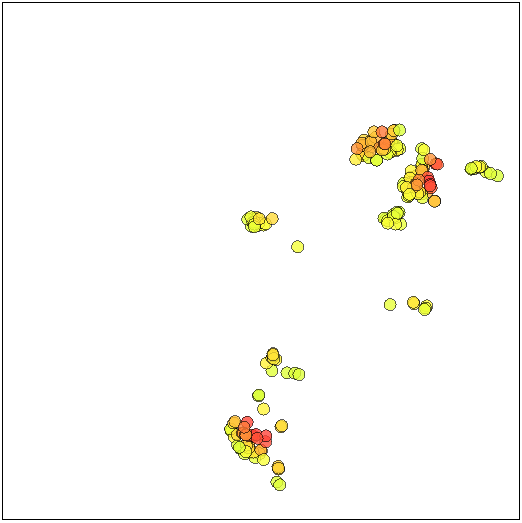}\\
  \rotatebox[origin=c]{90}{\footnotesize P.cysts} &
  \rotatebox[origin=c]{90}{\footnotesize $\tau=0.5$} &
  \cincludegraphics[width=.2\linewidth]{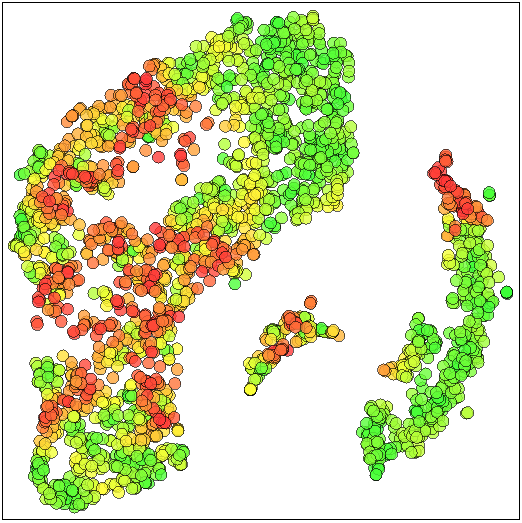} & \cincludegraphics[width=.2\linewidth]{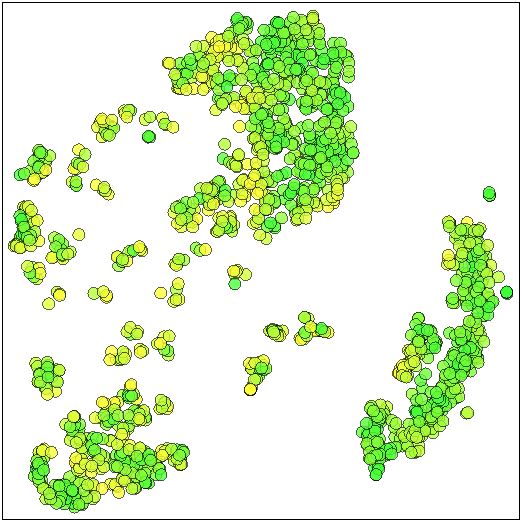} & \cincludegraphics[width=.2\linewidth]{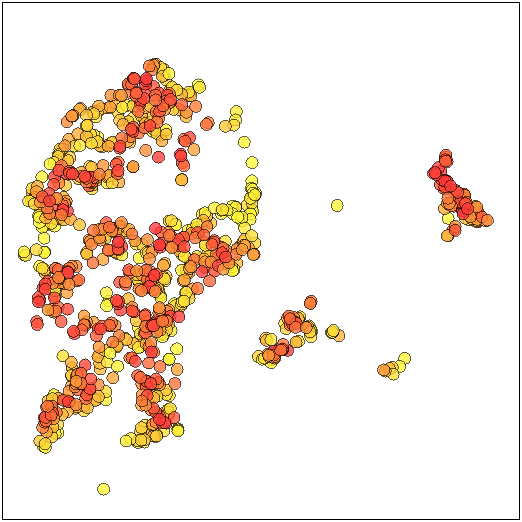}\\
  \rotatebox[origin=c]{90}{\footnotesize H.Larvae (I)} &
  \rotatebox[origin=c]{90}{\footnotesize $\tau=0.6$} &
  \cincludegraphics[width=.2\linewidth]{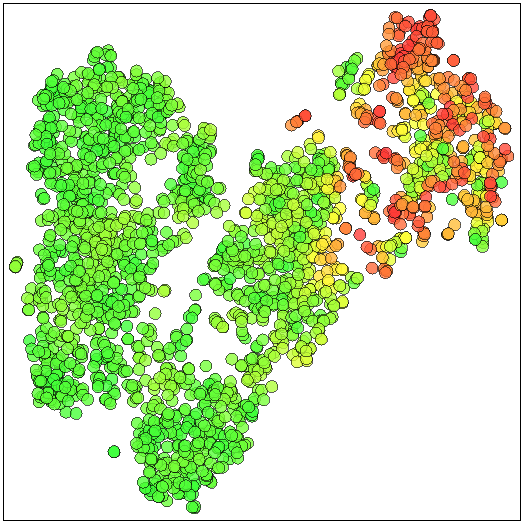} & \cincludegraphics[width=.2\linewidth]{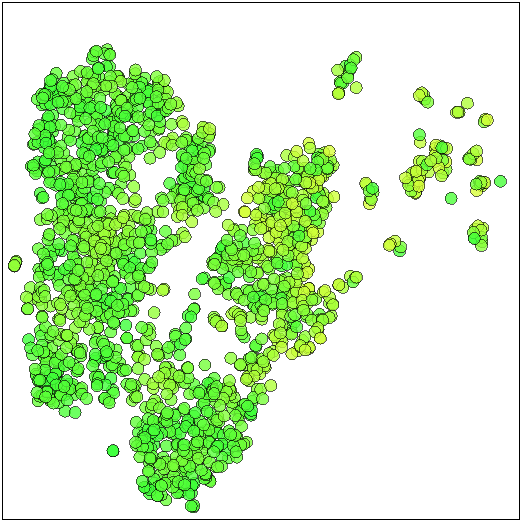} & \cincludegraphics[width=.2\linewidth]{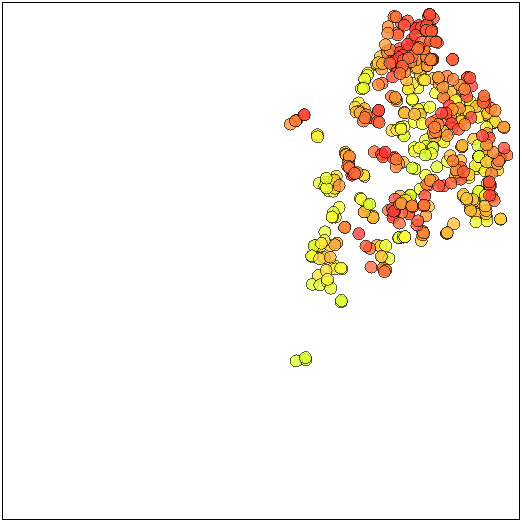}\\
  \rotatebox[origin=c]{90}{\footnotesize H.Eggs (I)} &
  \rotatebox[origin=c]{90}{\footnotesize $\tau=0.5$} &
  \cincludegraphics[width=.2\linewidth]{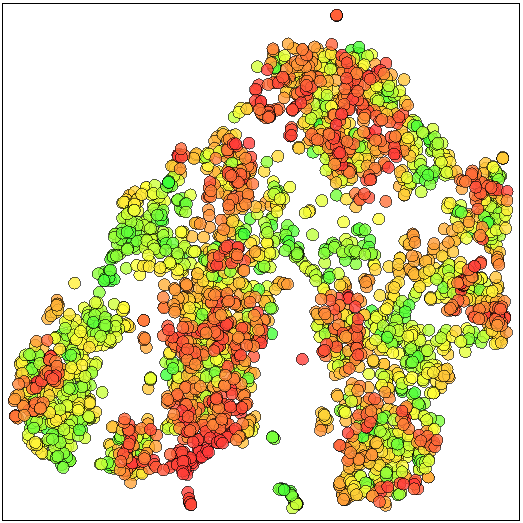} & \cincludegraphics[width=.2\linewidth]{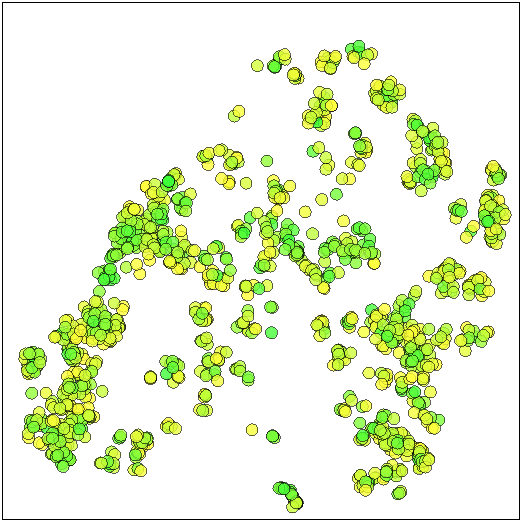} & \cincludegraphics[width=.2\linewidth]{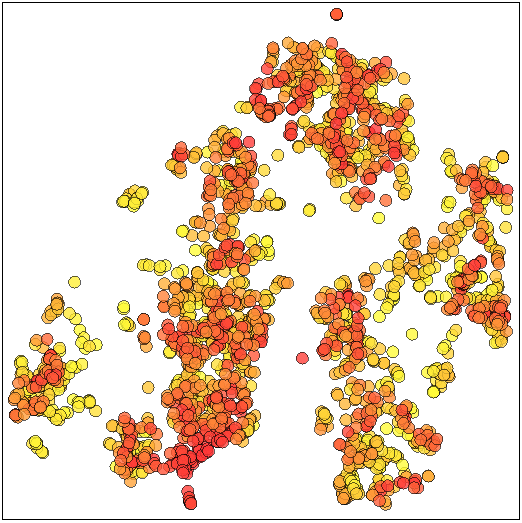}\\
  \rotatebox[origin=c]{90}{\footnotesize P.cysts (I)} &
  \rotatebox[origin=c]{90}{\shortstack[l]{\footnotesize $\tau=0.5$}} &
  \cincludegraphics[width=.2\linewidth]{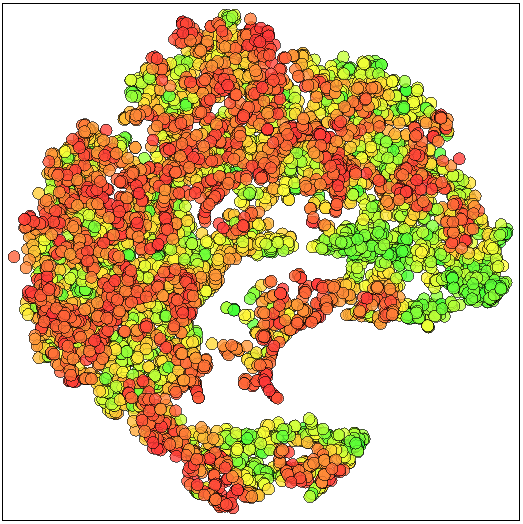} & \cincludegraphics[width=.2\linewidth]{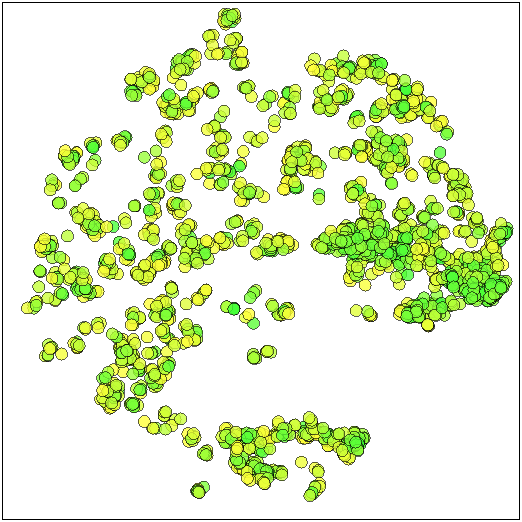} & \cincludegraphics[width=.2\linewidth]{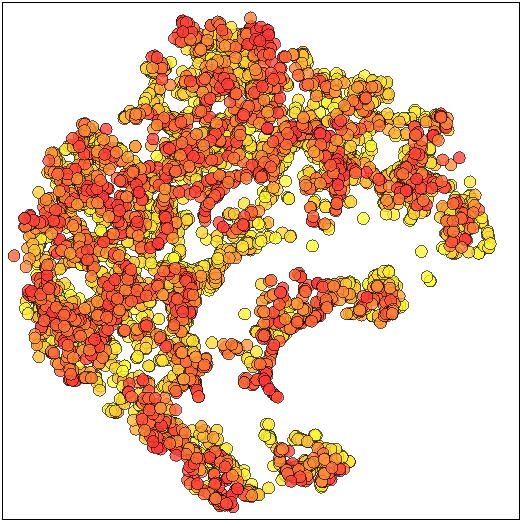}\\
\end{tabular}
\vspace{-0.2cm}
  \caption{Projections colored by label confidence (red=low confident, green=high confident). Rows are datasets (easiest at top, hardest at bottom). Columns show the entire set of supervised-and-unsupervised samples $S \cup U$, the high-confidence samples $L_c$ labeled by ALP, and the low-confidence samples $U \setminus L_c$ that go to manual labeling.}
  \label{f.mapmnist}
\end{figure}

As stated in Sec.~\ref{s.semi_sup}, users need to choose the threshold $\tau$ to specify which automatically-propagated labels they want to keep and which they wish to `override' manually. Figure~\ref{f.mapmnist} show the projections of all, respectively the most-confident samples selected by the user, for the six studied datasets. We see that the threshold $\tau$ varies relatively little (being either $0.5$ or $0.6$) across datasets. This indicates that a good default value to start with is $\tau=0.5$, after which users can tune $\tau$ upwards or downwards depending on the actual distribution of confidences in the projection. Overall, we can see that the more challenging is the dataset, the higher is the threshold $\tau$.

\subsubsection{Best label propagation approach}
\label{sec:q3}
\begin{figure}[htbp!]
\newcommand\cincludegraphics[2][]{\raisebox{-0.3\height}{\includegraphics[#1]{#2}}}
\centering
\vspace{-1.0cm}
\setlength{\tabcolsep}{1.2pt}
\begin{adjustbox}{width=\textwidth}
\begin{tabular}{cc|cccccccc}    
   & & \tiny \textbf{ILP} & \tiny \textbf{SALP} &  \multicolumn{1}{c}{\begin{tabular}[c]{@{}c@{}}\tiny \textbf{Labeled} \\ \tiny \textbf{(OPF-Semi)}\end{tabular}} & \multicolumn{1}{c}{\begin{tabular}[c]{@{}c@{}}\tiny \textbf{Labeled} \\ \tiny  \textbf{(OPF-Semi+user)}\end{tabular}}  & \tiny $|U|$ & \tiny $|L_c|$ & \tiny $|U\setminus L_c|$ & \tiny $|L_i|$  \\
  \hline
  \rotatebox{90}{\tiny MNIST} &
  \rotatebox{90}{\tiny $\tau=0.5$} &
  \cincludegraphics[width=.18\linewidth]{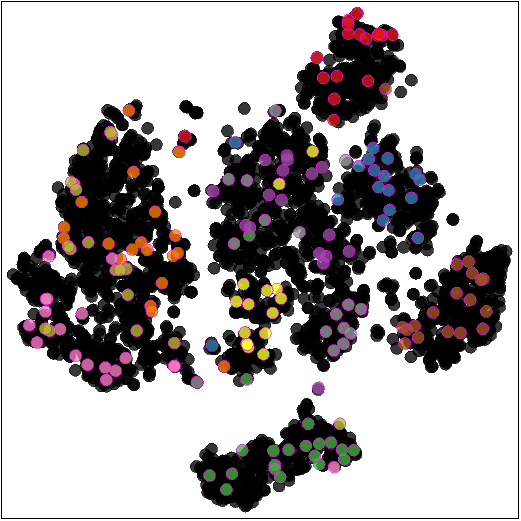} &
  \cincludegraphics[width=.18\linewidth]{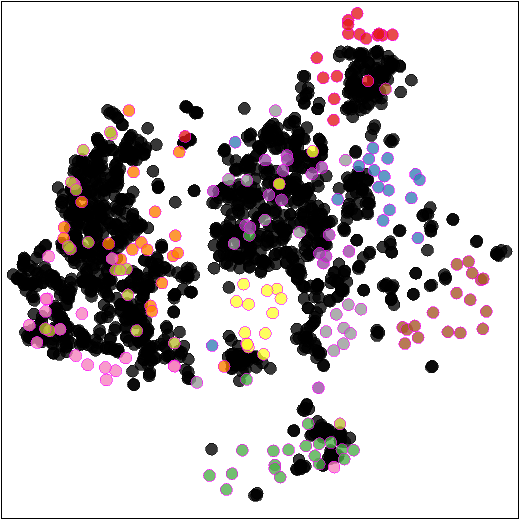} &
  \cincludegraphics[width=.18\linewidth]{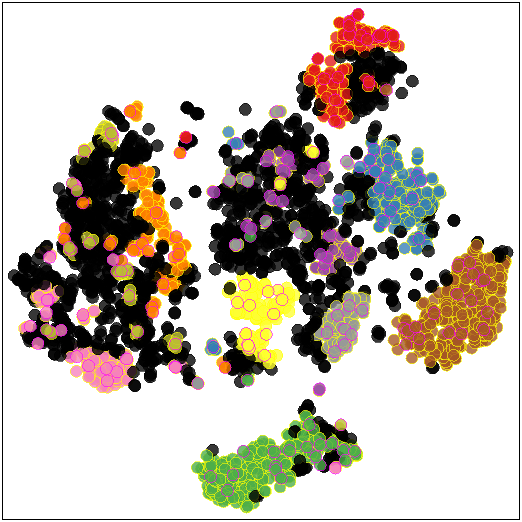} & \cincludegraphics[width=.18\linewidth]{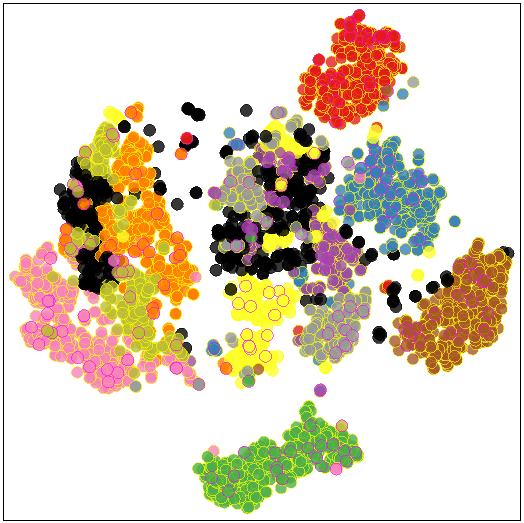} &
  \tiny 3325 &
  \tiny 1690 &
  \tiny 1635 &
  \tiny 1182 \\
  \rotatebox{90}{\tiny H.Eggs} &
  \rotatebox{90}{\tiny $\tau=0.6$} &
  \cincludegraphics[width=.18\linewidth]{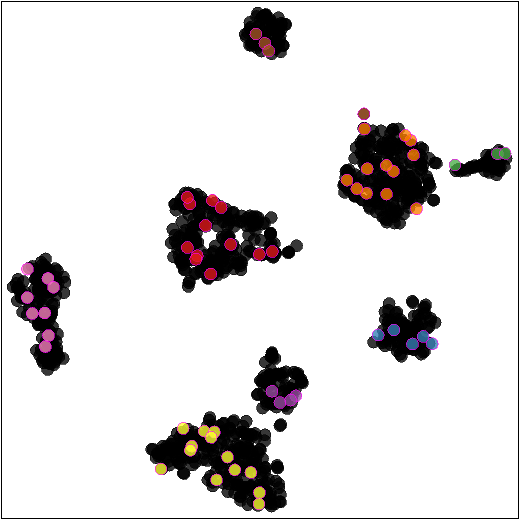} & \cincludegraphics[width=.18\linewidth]{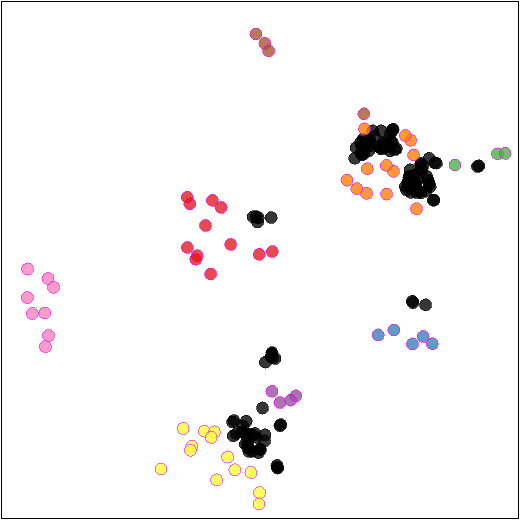} & \cincludegraphics[width=.18\linewidth]{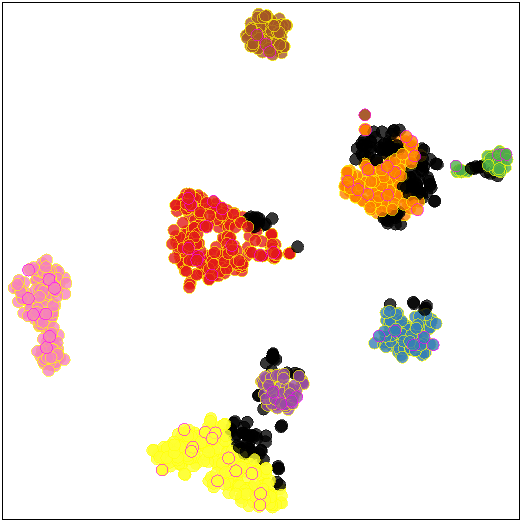} & \cincludegraphics[width=.18\linewidth]{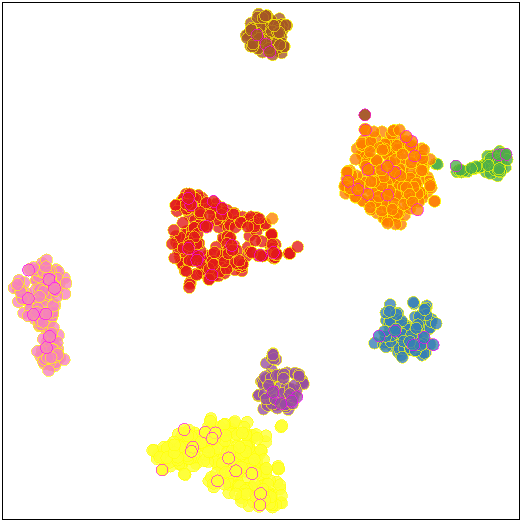} &
  \tiny 1176 &
  \tiny 1022 &
  \tiny 154 &
  \tiny 154 \\
  \rotatebox{90}{\tiny P.cysts} &
  \rotatebox{90}{\tiny $\tau=0.5$} &
  \cincludegraphics[width=.18\linewidth]{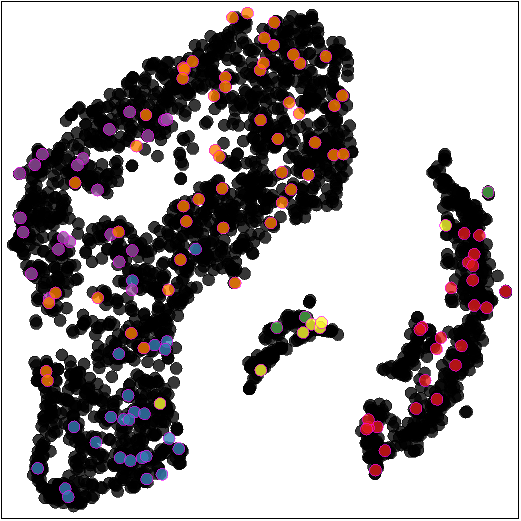} & \cincludegraphics[width=.18\linewidth]{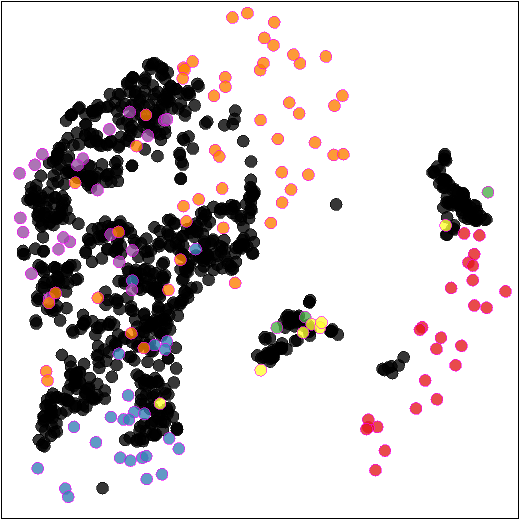} & \cincludegraphics[width=.18\linewidth]{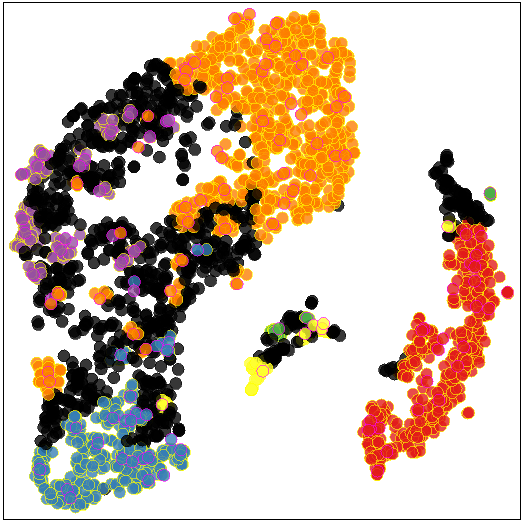} & \cincludegraphics[width=.18\linewidth]{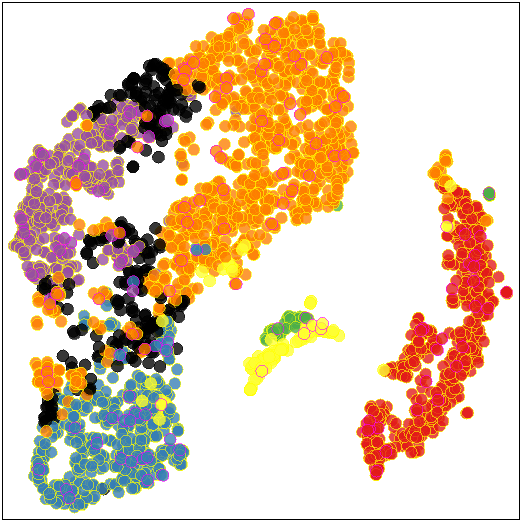} &
  \tiny 2562 &
  \tiny 1643 &
  \tiny 919 &
  \tiny 666 \\
  \rotatebox{90}{\tiny H.Larvae (I)} &
  \rotatebox{90}{\tiny $\tau=0.6$} &
  \cincludegraphics[width=.18\linewidth]{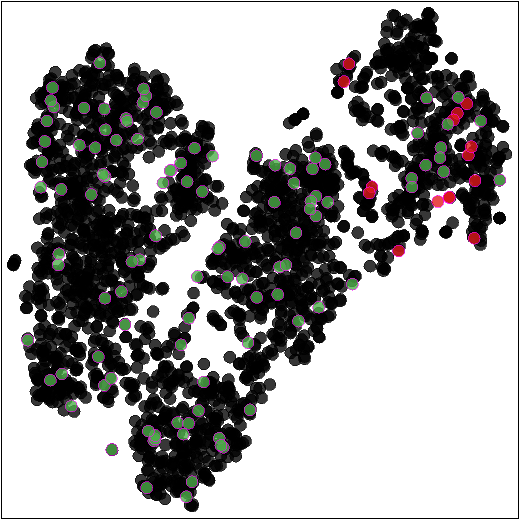} & \cincludegraphics[width=.18\linewidth]{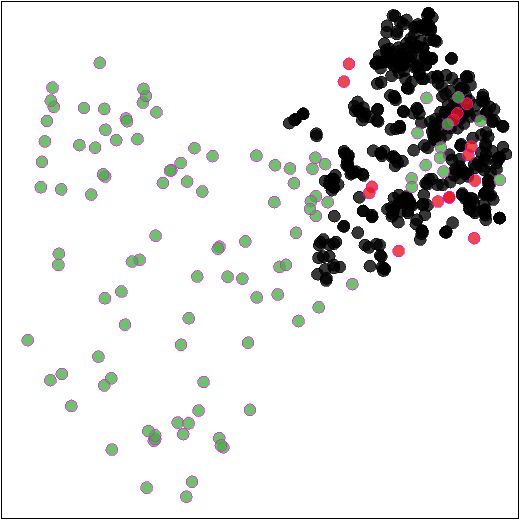} & \cincludegraphics[width=.18\linewidth]{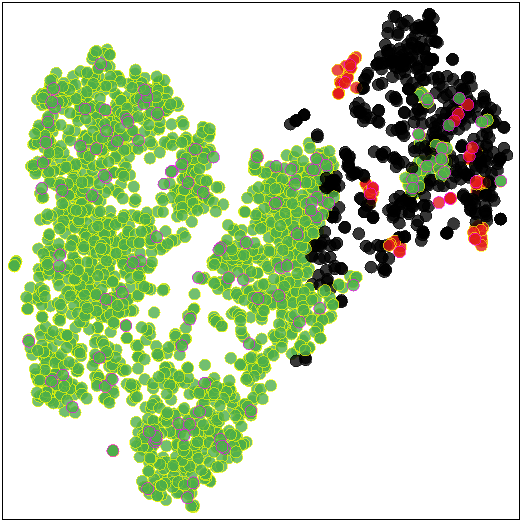} & \cincludegraphics[width=.18\linewidth]{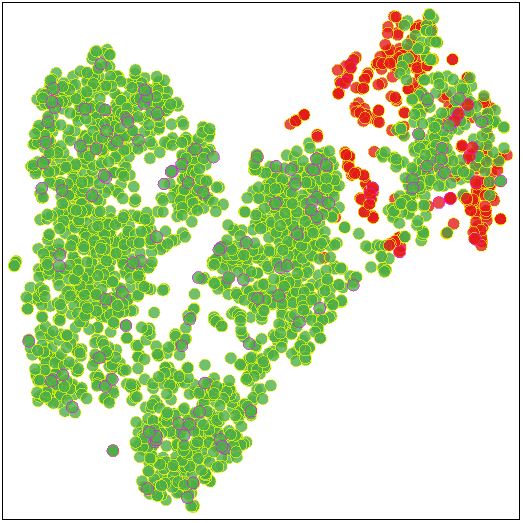} &
  \tiny 2337 &
  \tiny 1813 &
  \tiny 524 &
  \tiny 524 \\
  \rotatebox{90}{\tiny H.Eggs (I)} &
  \rotatebox{90}{\tiny $\tau=0.5$} &
  \cincludegraphics[width=.18\linewidth]{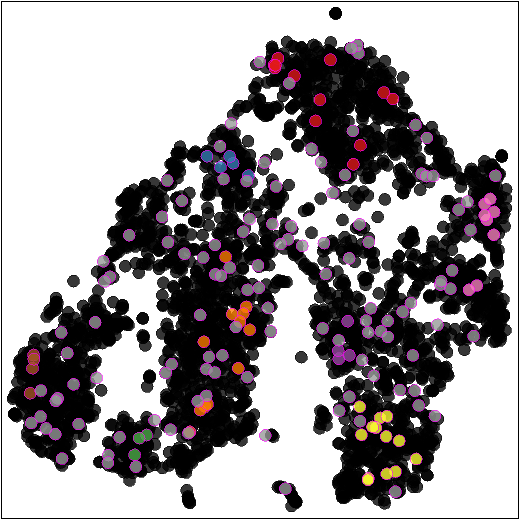} & \cincludegraphics[width=.18\linewidth]{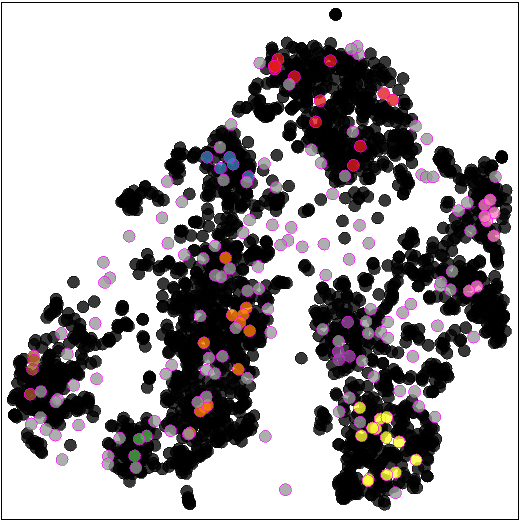} & \cincludegraphics[width=.18\linewidth]{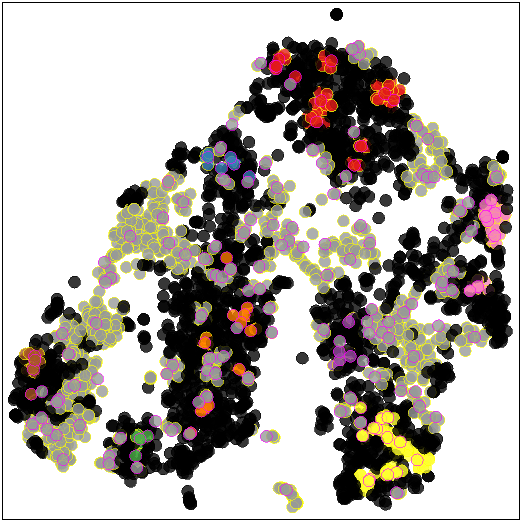} & \cincludegraphics[width=.18\linewidth]{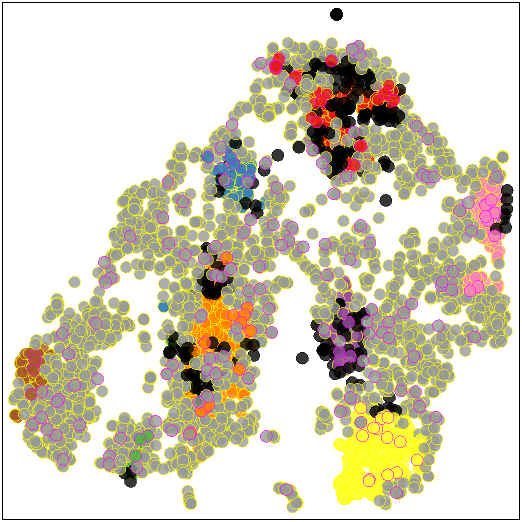} &
  \tiny 3400 &
  \tiny 983 &
  \tiny 2417 &
  \tiny 2076 \\
  \rotatebox{90}{\tiny P.cysts (I)} &
  \rotatebox{90}{\tiny $\tau=0.5$} &
  \cincludegraphics[width=.18\linewidth]{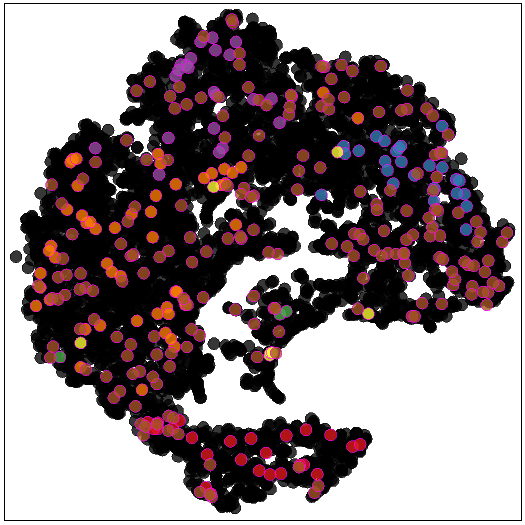} & \cincludegraphics[width=.18\linewidth]{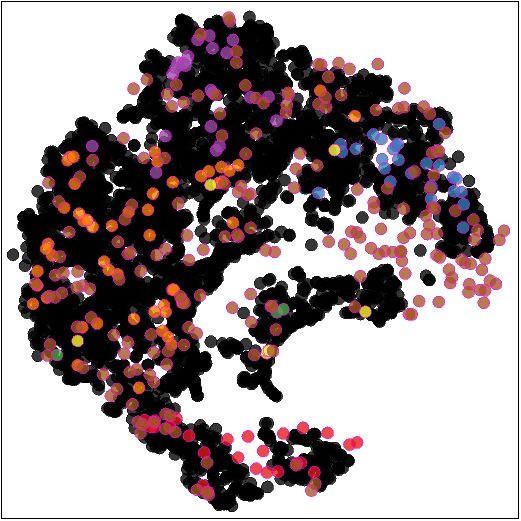} & \cincludegraphics[width=.18\linewidth]{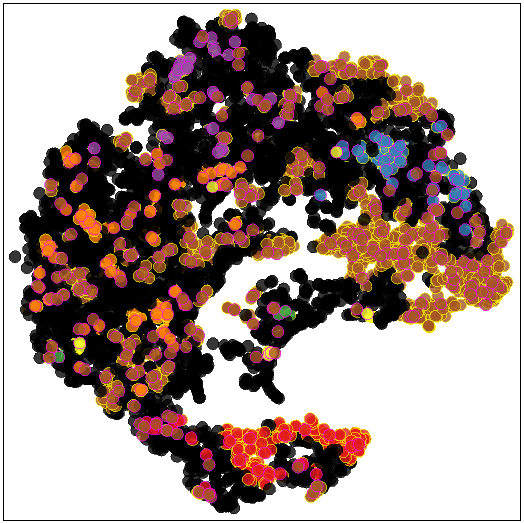} & \cincludegraphics[width=.18\linewidth]{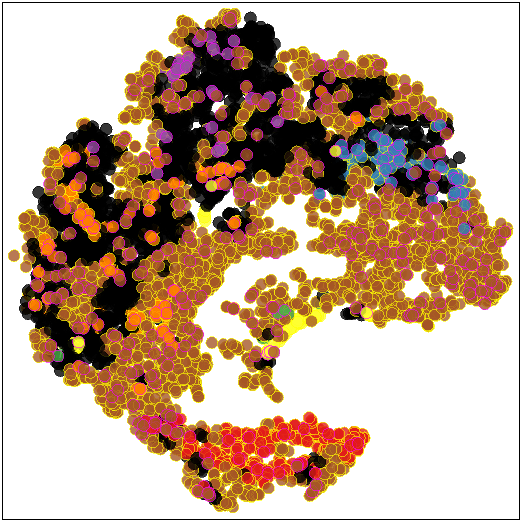} &
  \tiny 6363 &
  \tiny 2215 &
  \tiny 4148 &
  \tiny 1733 \\
  \end{tabular}
\end{adjustbox}
\vspace{-0.2cm}
  \caption{Comparison of different label propagation methods (columns) for different datasets (rows). From left to right: ILP, SALP, labels automatically propagated by OPF-Semi, and final labeling result of SALP together with OPF-Semi. Colors indicate labels given by either supervised samples (ILP, SALP) or both unsupervised and propagated labels (OPF-Semi, OPF-semi+user)). Black shows samples to be considered by manual propagation (three left columns), and samples skipped by manual propagation (right column). Sample set sizes are shown to the right.}
  \label{f.mnist}
\end{figure}

 Table~\ref{t.results1} showed that OPF-Semi 2D is the winner for automatic label propagation (ALP). Hence, the next question is how well this method would compare against interactive label propagation (ILP)\,\cite{BenatoSibgrapi:2018}, which uses manual label propagation to \emph{all} unsupervised labels, and our new semi-automatic label propagation (SALP), which uses manual label propagation to samples with low-confidence unsupervised labels only.
Figure~\ref{f.mnist} illustrates the ILP and SALP projections for the studied datasets. A key advantage of SALP over ILP is that it shows only the least confident samples (according to OPF-Semi 2D) to the user, hence reducing the effort needed to understand the picture (and also reducing clutter and overlap in the projection), thus making the interactive labeling task easier. We discuss next several observations relating ILP to SALP in Fig.~\ref{f.mnist}, as well as observations we made during the actual interactive labeling process. 

For the MNIST dataset, the user propagated labels to $1864$ unsupervised samples on average (over the three considered runs) when using ILP. When using SALP, this number dropped to $1182$ samples. This pattern of less effort for SALP is consistent over all other datasets, as discussed next.

For the H.Eggs dataset, we see that the ILP projection shows well-separated sample groups from distinct classes (colors). This indicates that separating classes in feature space is relatively easy. This is confirmed in turn by the fact that we only have very few low-confidence samples after running OPF-Semi 2D (black dots in the SALP projection). Hence, while labeling in ILP can proceed very easily, given the good cluster separation, labeling in SALP is \emph{even easier}, since we have both good cluster separation \emph{and} a low number of samples to label. In this case, the user propagated labels to $1171$ samples in ILP and to only $154$ samples in SALP.  

For P.cysts, the projections a less clear visual separation of same-class (same color) points in groups. This makes interactive label propagation more challenging for both ILP and SALP. The user propagated labels to $1999$ samples in ILP and to $919$ samples in SALP. For SALP, we see that OPF-Semi 2D propagated labels in more central regions of the visible groups where, hence, confidence is high. The remaining confusion regions (black points) are solved by the user.

For. H.Larvae, we notice that supervised impurity samples (green) are all over the projection, whereas the supervised H.Larvae samples (red) are more concentrated in the top-right of the projection. Given this quite good visual separation, propagating the impurity label using ILP is relatively easy for most parts of the projection. However, this still takes manual effort. Using SALP, such `easy' areas are solved automatically, and the user is left with only the more difficult region at the top-right, where green meets red, to solve. In ILP, the user propagated labels for $2080$ samples on average while in SALP this number was $524$ samples. 

For H.Eggs dataset with impurity, the supervised impurity samples (gray) fall between groups of colored points (actual H.Eggs classes) in the projection. 
In contrast to the earlier datasets, we see many more black points in SALP, meaning that OPF-Semi 2D has difficulties in automatically propagating labels. This matches the fact that datasets with impurities are considerably harder. 
For this dataset, the user propagated labels to more points in SALP ($2076$) than ILP ($1787$). This seems to support the evidence that the simplification of the SALP projection by removing high-confidence points, even though minor in this case, was enough to help the user see more structure in the projection along which she could propagate labels. Also, as for P.cysts, we see that OPF-Semi 2D propagates labels in more central regions of the visible groups, leaving the rest to the user.

Finally, for P.cysts with impurities, the supervised impurity samples (brown) are spread out over the entire projection. The supervised P.cysts samples (other colors than brown) are mixed quite strongly, and the projection shows little structure -- roughly, one large and one small crescent-shaped group. This is the most challenging dataset for manual label propagation and classification among the evaluated datasets. This difficulty can be noted by comparing P.cysts and H.Eggs both without impurities. For P.cysts, even without impurities, the classes are mixed in the projection. However, the classes are well separated in the projection for the H.Eggs dataset without impurities. When adding the impurities to those datasets, the difficulty increases for the classifiers, as shown in  Sec.
~\ref{sec:q4}.

As for H.Eggs, OPF-Semi 2D finds only few confident samples, so the manual labeling effort is quite similar for both ILP and SALP. This is matched by the actual number of points to which the user actually propagated labels ($1787$ with ILP \emph{vs} $1733$ with SALP). Even though these figures are almost identical, the main benefit for SALP here is that OPF-Semi 2D already filtered the easy cases (high confidence) points, thereby focusing the user's effort to the more difficult cases.

\subsubsection{End-to-end value of SALP}
\label{sec:q4}
We have seen that SALP decreases the user's effort in label propagation. A final question we answer is: How much added-value does SALP bring, in terms of classification quality, as opposed to the earlier similar method, ILP, or to the best fully-automatic counterpart we found, OPF-Semi 2D? Table~\ref{t.results2} answers this by showing the average and stardard deviation of $\kappa$ on the test set $T$ for each considered dataset. The table further shows the sizes of $S$, $U$, and $S\cup U$, and the mean accuracy values in label propagation for OPF-Semi 2D, ILP, and SALP. It is important to highlight that the propagation accuracy for SALP considers not only the low-confident samples labeled by the user, but the high-confident ones automatically treated by OPF-Semi 2D. We see that SALP consistently obtained the best classification results on unseen $T$ for all datasets. This proves that SALP is, indeed, of added value with respect to earlier existing methods -- using it yields better classifiers in the end. Separately, we see that, for all but the simplest datasets (MNIST and H.Eggs), SALP also yields the best label propagation accuracy.

\input{results_2_new}

\subsubsection{How do results depend on projection quality}
\label{sec:q5}
We did the same experiments discussed in the sections so far using UMAP\,\cite{UMAP:2018} instead of t-SNE as a projection technique. Overall, we noticed worse results, in terms of label propagation accuracy and classifier quality ($\kappa$) than when using t-SNE. This indicates that the neighborhood preservation quality of a projection (which is higher for t-SNE than for UMAP) is am important factor for out method. Note also that the trends observed so far linking obtained SALP and ILP quality with the dataset size and difficulty cannot be ascribed to us having used `optimal' projections by a lucky setting of the projection-method parameters: Indeed, both UMAP and t-SNE are non-deterministic methods. 

\section{Discussion}
\label{s.discussion}
We next discuss several aspects of our method
\subsection{Using the $n$D \emph{vs} 2D feature space}
An interesting question is how the fully automatic label propagation (ALP) performs when using the latent $n$D feature space \emph{vs} the 2D projection space.  
Figure~\ref{f.2d_nd_feat} shows the average $\kappa$ classification values for LapSVM and OPF-Semi using these two spaces for the OPF and SVM classifiers respectively. Datasets are sorted along the $x$ axis by decreasing order of the $\kappa$ value for OPF-Semi 2D. We see that LapSVM leads to better results in 2D than in $n$D for half of the datasets, while OPF-Semi does that for \emph{all} datasets.
This essentially tells that the 2D projection space, created by t-SNE, is able to retain all needed information to enable the desired label propagation and, next, good-quality classifier construction. This is an important result, as it justifies next presenting the 2D projection space to the user as the sole information based on which she will perform the manual label propagation. We also see that the trend of the $\kappa$ values along the $x$ axis, for both the 2D and $n$D variants, matches the perceived difficulty of the datasets: High $\kappa$ values correspond to easier datasets (left), while lower $\kappa$ values correspond to the harder datasets with impurities (to the right). Finally, we plot here also the $\kappa$ values for ILP and SALP (curves in the figures). In all cases, these curves are above the automatic methods, showing that adding manual effort pays off. The SALP curve is above the ILP one, showing that the optimal design is reached by \emph{combining} automatic and manual label propagation (both executed in the 2D space).

\input{graphs/nd_2d}

\subsection{User effort reduction}
Besides achieving the best classification results, as compared to both fully-automatic and fully-manual (ILP) label propagation, SALP also reduces the \emph{manual} effort as compared to ILP. 
Figure~\ref{f.user_effort} shows this by depicting the percentage of samples labeled by the user over total number of samples to label ($|U|$) per dataset and for ILP and SALP. For SALP, this measurement excludes, indeed, the automatically-labeled samples by OPF-Semi 2D. Datasets are sorted along $x$ by increasing $|U|$, i.e, from the smallest to the largest dataset. 
Figure~\ref{f.user_effort} reveals several insights. First, assuming that the labeling effort is proportional with the number of labeled samples and the effort per sample is the same for ILP and SALP (which should be the case given that the two methods share the same visualization and interaction), we see that the ILP effort is always larger than the SALP effort, except for H.Eggs with impurities. Secondly, the percentage of propagated samples for ILP decreases with the dataset size. This can be explained by the difficulty of propagating labels in projections showing many points, where overlap and clutter become issues. We note an opposite for trend SALP: The percentage of propagated samples increases with dataset size. The trend breaks for the largest dataset (Prot.c.(I), $6363$ samples), about twice larger than the second-largest dataset (H.Eggs(I), $3400$ samples). Here, the projection is likely quite dense and cluttered, so manual propagation becomes similarly hard for ILP and SALP.

In parallel, we observe that the number of samples $U\setminus L_c$, those above the threshold $\tau$ and low-confidence labels to OPF-Semi, also increases with the dataset size. Thus, the amount of samples $U\setminus L_c$ presented to the user to propagate labels with SALP increases with dataset size. One case in point is the H.Eggs with impurities dataset. This dataset has the largest percentage of annotated samples by SALP, exceeding also ILP.  This is explained by the size of the dataset (second largest one) and the fact that its projection makes it reasonably easy to propagate labels for the large impurity class (Fig.~\ref{f.mapmnist}).


\input{graphs/user_effort.tex}
\input{graphs/effectiveness.tex}

\subsection{Effectiveness}
As shown in Fig.~\ref{f.2d_nd_feat}, SALP consistently yields best classification results, for both SVM and OPF classifiers, overpassing fully manual propagation methods (ILP) and the best fully automatic one (OPF-Semi 2D). The gains of SALP are higher for the more challenging datasets, where fully automatic methods encounter challenges. Conversely, where such methods work well, they reduce user effort as compared to fully manual propagation (ILP). In brief, this shows that the combination of automatic methods with human insights is indeed of added value both in increasing classifier quality and decreasing the effort needed to achieve it. 

It is next interesting to compare the \emph{normalized gain} of ILP \emph{vs} SALP. We define this as the obtained $\kappa$ value (what we get) divided by the percentage of manually labeled samples (what we need to pay). Figure~\ref{f.effectiveness} shows this normalized gain for ILP and SALP for both SVM and OPF classifiers. We see that SALP has far larger normalized gains than SALP for smaller datasets, while differences become quite small for the two largest datasets.

\subsection{Manual sample selection justification}

In classical pipelines, expert users would label samples in an empirical order. In pipelines that consider active learning methods, the sample informativeness can be used to suggest samples in each iteration for user supervision. However, those approaches do not usually explore the ability of humans in abstracting information from data visualization. Given that their labeling effort is limited (and their cost is high), the aim is to maximize the `added value' of creating extra labels manually. Our hypothesis (which we show, by our experiments, to hold) is that, when expert users are offered hints in terms of sample similarity (via the 2D projection and its tooltips) and by the confidence of an automatic labeler (color-coded in the projection), they can manually create extra labels that have a higher added-value (for classification accuracy) than fully automatic methods can achieve. 

The core point of manual labeling is to enable users with expert knowledge select the samples they think are most relevant for constructing a good training set. Answering the question of why expert users would select a certain sample subset rather than another one is not something we can argue theoretically, as it depends on a multitude of factors -- first and foremost, the training of the expert and how this training determines the expert to consider a given image more (or less) relevant for being labeled in a certain way. 

\subsection{Limitations}
Several limitations exist to our approach, as follows. First, \emph{validation} is limited to six datasets, two classifier techniques, and one user performing manual labeling. Measuring the added-value of SALP for more (dataset, classifier, user) combinations would bring more insights into the effectiveness of the method. Secondly, while the added-value of the 2D t-SNE projection space in capturing information needed for good label propagation has been demonstrated both for automatic methods and manual ones, the actual effect of t-SNE's distortions has not been quantitatively gauged. Using projection accuracy metrics such as stress, trustworthiness, continuity, or neighborhood hit\,\cite{nonato18} can be used to find such correlations. On the other hand, using visual tools\,\cite{nonato18} that highlight such errors in specific projection areas can help the user to achieve more accurate and/or faster manual label propagation.

\section{Conclusion}
\label{s.conclusion}
We proposed a combined automatic-and-user-driven approach for creating labeled samples for sparsely-annotated datasets for the purpose of training classifier models. For this, we extract dataset features using Autoencoder Neural Networks and next reduce these to a 2D space using t-SNE. We next automatically propagate labels from the (few) supervised to unsupervised samples in this 2D space, while monitoring the propagation confidence. For low confidence labeled samples, we allow the user to manually annotate them by using the visual insights encoded in the 2D projection annotated with the supervised sample labels. Several quantitative results follow: First, we showed that the 2D projection space leads to higher-accuracy automatic label propagation than the high-dimensional latent space extracted by the autoencoder. To our knowledge, this insight is new, and suggests new ways for dimensionality reduction. Secondly, we show that our semi-supervised method, combining the OPF-Semi automatic label propagation with user-driven manual label propagation, both done in the 2D space, achieves higher classification quality than both fully-automatic and fully-manual label propagation. This opens the way to different methods for combining automatic and human-centered methods for the engineering of high-quality machine learning systems.

Future work will consider the use of the proposed semi-automatic label propagation method in Active Learning (AL) scenarios. We expect that AL looping can improve classification results as long as the propagation accuracy increases. Also, we intend to consider metric learning approaches that might improve the 2D projection of the feature space. We are interested in methods that allow the comparison between training and testing data. Specifically, we intend to investigate methods such as the exemplar-centered High Order Parametric Embedding~\cite{Min:2017}. Separately, we plan to perform more extensive validation studies measuring the added-value of our approach for more types of datasets, classification methods, and using additional visual analytics techniques to help users to propagate labels better and faster.


\section*{Acknowledgments}
The authors are grateful to FAPESP grants \#2014/12236-1, \#2016/25776-0 and \#2017/25327-3, and CNPq grants 303808/2018-7. The views expressed are those of the authors and do not reflect the official policy or position of the S\~ao Paulo Research Foundation.



  \bibliographystyle{elsarticle-num} 
  \bibliography{refs}





\end{document}

%% file: split_data.tex
\renewcommand{\arraystretch}{0.7}
\begin{table}[h]
\footnotesize
\centering
\caption{Number of samples in $S$, $U$ and $T$ for each dataset: MNIST, Parasites, and Parasites with impurity (I).}
\label{t.split_data}
\begin{tabular}{lcccc}
\multicolumn{1}{c}{Dataset} & $|S|$ & $|U|$ & $|S\cup U|$ & $|T|$ \\
\hline
MNIST   &    175    &    3325   &    3500   &    1500   \\ 
H.Eggs      &    61     &    1176   &   1237    &    531    \\ 
P.cysts     &    134    &    2562   &    2696   &    1156   \\ 
H.Larvae (I)    &    122    &    2337   &    2459   &   1055    \\ 
H.Eggs (I)  &    178    &    3400   &    3578   &    1534   \\ 
P.cysts (I)     &    334    &   6363    &    6697   &    2871   \\
\hline
\end{tabular}
\end{table}

%% file: results_1_new.tex
\renewcommand{\arraystretch}{0.7}
\begin{table*}[h]
\footnotesize
\centering
\caption{Average $\kappa$ and its standard deviation for SVM and OPF classifiers on the $T$ set for MNIST, Parasites without impurity, and Parasites with impurity (I). The best results for each dataset are in bold.}
\label{t.results1}
\begin{adjustbox}{width=\textwidth}
\begin{tabular}{lllrrrrrrrr}
\multicolumn{2}{c}{\textbf{Dataset}} & \multicolumn{1}{c}{\begin{tabular}[c]{@{}c@{}}\textbf{Propagation} \\ \textbf{Technique}\end{tabular}} & \multicolumn{1}{c}{\begin{tabular}[c]{@{}c@{}}$\mid S \mid$\end{tabular}} & \multicolumn{1}{c}{\begin{tabular}[c]{@{}c@{}}$\mid U \mid$\end{tabular}} & \multicolumn{1}{c}{\begin{tabular}[c]{@{}c@{}}\textbf{Average}\\ \textbf{Propagation}\\ \textbf{Accuracy}\end{tabular}} & \multicolumn{1}{c}{\begin{tabular}[c]{@{}c@{}} $\mid S\cup U \mid$\end{tabular}} & \multicolumn{1}{c}{\begin{tabular}[c]{@{}c@{}}\textbf{Average} $\kappa$ \\ \textbf{(SVM)}\end{tabular}} & \multicolumn{1}{c}{\begin{tabular}[c]{@{}c@{}}\textbf{Average} $\kappa$ \\ \textbf{(OPF)}\end{tabular}}  \\
\hline
\multicolumn{2}{l}{\multirow{5}{*}{MNIST}}  &   No label prop.  &   175 &   -   &   -   &   175 &   \textbf{0.813415 $\pm$ 0.001}   &   0.709450 $\pm$ 0.021 \\
\multicolumn{2}{l}{}    &    LapSVM ($n$D)   &   175     &   3325    &   0.095639    &   3500    &   0.000000 $\pm$ 0.000    &   0.051110 $\pm$ 0.006     \\
\multicolumn{2}{l}{}    &    OPF-Semi ($n$D)  &   175     &   3325    &   0.763308    &   3500    &   0.736685 $\pm$ 0.053    &   0.716913 $\pm$ 0.048   \\
\multicolumn{2}{l}{}    &   LapSVM (2D)     &   175     &   3325    &   0.574236    &   3500    &   0.521970 $\pm$ 0.065    &   0.580445 $\pm$ 0.047   \\ 
\multicolumn{2}{l}{}    &   OPF-Semi (2D)   &   175     &   3325    &   \textbf{0.796592}   &   3500    &   0.780197 $\pm$ 0.008    &     \textbf{0.751794 $\pm$ 0.010}    \\
\hline
\multicolumn{2}{l}{\multirow{5}{*}{H.Eggs}}     &   No label prop.  &   61  &   -   &   -   &   61  &   0.961366 $\pm$ 0.023    &   0.941358 $\pm$ 0.026     \\
\multicolumn{2}{l}{}    &   LapSVM ($n$D)     &    61     &    1176   &    0.886338   &    1236   &    0.873472 $\pm$ 0.035   &    0.877344 $\pm$ 0.037   \\
\multicolumn{2}{l}{}    &    OPF-Semi ($n$D)  &    61     &    1176   &    0.947563   &   1236    &   0.938317 $\pm$ 0.049    &   0.938323 $\pm$ 0.051    \\
\multicolumn{2}{l}{}    &    LapSVM (2D)    &   61  &   1176    &   0.768141    &   1236    &   0.722691 $\pm$ 0.041    &   0.727673 $\pm$ 0.043     \\ 
\multicolumn{2}{l}{}    &   OPF-Semi (2D)   &   61  &   1176    &   \textbf{0.982993}   &   1236    &    \textbf{0.983621 $\pm$ 0.011}  &     \textbf{0.982146 $\pm$ 0.008}   \\
\hline
\multicolumn{2}{l}{\multirow{5}{*}{P.cysts}}    &   No label prop.  &   134     &    -  &   -   &   134     &   \textbf{0.823106 $\pm$ 0.016}   & 0.762682 $\pm$ 0.008    \\
\multicolumn{2}{l}{}    &   LapSVM ($n$D)     &   134     &   2562    &   0.521598    &   2696    &   0.346761 $\pm$ 0.001    &   0.371770 $\pm$ 0.005   \\
\multicolumn{2}{l}{}    &   OPF-Semi ($n$D)   &   134     &   2562    &   0.802238    &   2696    &   0.740287 $\pm$ 0.036    &   0.724182 $\pm$ 0.053   \\
\multicolumn{2}{l}{}    &   LapSVM (2D)     &   134     &   2562    &   0.787666    &   2696    &   0.597541 $\pm$ 0.212    &   0.592956 $\pm$ 0.187   \\
\multicolumn{2}{l}{}    &   OPF-Semi (2D)   &   134     &   2562    &   \textbf{0.838017}   &   2696    &   0.801953 $\pm$ 0.021    &    \textbf{0.786383 $\pm$ 0.022}   \\
\hline
\multicolumn{2}{l}{\multirow{5}{*}{H.Larvae (I)}}   &   No label prop.  & 122   &   -   &   -   &   122     &   0.375378 $\pm$ 0.333    &    0.531080 $\pm$ 0.035    \\
\multicolumn{2}{l}{}    &   LapSVM ($n$D)     &   122     &   2337    &   0.882613    &   2459    &   0.121253 $\pm$ 0.086    &   0.173416 $\pm$ 0.088   \\
\multicolumn{2}{l}{}    &   OPF-Semi ($n$D)   &   122     &   2337    &   0.919075    &   2459    &   \textbf{0.601003 $\pm$ 0.073}   &     \textbf{0.602001 $\pm$ 0.066}   \\
\multicolumn{2}{l}{}    &   LapSVM (2D)     &   122     &   2337    &   0.127086    &   2459    &   0.000000 $\pm$ 0.000    &   0.008665 $\pm$ 0.005   \\ 
\multicolumn{2}{l}{}    &   OPF-Semi (2D)   &   122     &   2337    &   \textbf{0.924405}   &   2459    &   \textbf{0.569164 $\pm$ 0.070}   &     \textbf{0.556642 $\pm$ 0.059}  \\
\hline
\multicolumn{2}{l}{\multirow{5}{*}{H.Eggs (I)}}     &   No label prop.  &   178     &   -   &   -   &   178     &   \textbf{0.705972 $\pm$ 0.037}    &    \textbf{0.568304 $\pm$ 0.034}   \\
\multicolumn{2}{l}{}    &   LapSVM ($n$D)     &   178     &   3400    &   0.654118    &   3578    &   0.000000 $\pm$ 0.000    &   0.076043 $\pm$ 0.016   \\
\multicolumn{2}{l}{}    &   OPF-Semi ($n$D)   &   178     &   3400    &   0.504510    &   3578    &   0.373763 $\pm$ 0.029    &   0.391894 $\pm$ 0.021   \\
\multicolumn{2}{l}{}    &   LapSVM (2D)     &   178     &   3400    &   0.146274    &   3578    &   0.086608 $\pm$ 0.031    &   0.109734 $\pm$ 0.035   \\ 
\multicolumn{2}{l}{}    &   OPF-Semi (2D)   &   178     &   3400    &   \textbf{0.729608}   &   3578    &   0.611144 $\pm$ 0.071    &    \textbf{0.544552 $\pm$ 0.047}   \\
\hline
\multicolumn{2}{l}{\multirow{5}{*}{P.cysts (I)}}    &   No label prop.  &   334     &   -   &   -   &   334     &   \textbf{0.628584 $\pm$ 0.024}     &    \textbf{0.476051 $\pm$ 0.010}   \\
\multicolumn{2}{l}{}    &    LapSVM ($n$D)    &    334    &    6363   &    \textbf{0.622662}  &   6697    &    0.232800 $\pm$ 0.104   &    0.202826 $\pm$ 0.030     \\
\multicolumn{2}{l}{}    &    OPF-Semi ($n$D)  &    334    &    6363   &    0.468804   &    6697   &    0.343356 $\pm$ 0.034   &    0.337168 $\pm$ 0.032   \\
\multicolumn{2}{l}{}    &    LapSVM (2D)    &    334    &    6363   &    0.128818   &    6697   &    0.045538 $\pm$ 0.038   &    0.079854 $\pm$ 0.024   \\
\multicolumn{2}{l}{}    &    OPF-Semi (2D)  &    334    &    6363   &    0.605427   &    6697   &    0.429731 $\pm$ 0.013   &    0.396645 $\pm$ 0.009   \\
\hline
\end{tabular}
\end{adjustbox}
\end{table*}

%% file: results_2_new.tex
\renewcommand{\arraystretch}{0.8}
\begin{table*}[!h]
\footnotesize
\centering
\caption{Average $\kappa$ and its standard deviation for the SVM and OPF classification results on unseen test data $T$ for the studied datasets. Best results per dataset are in bold.}
\label{t.results2}
\begin{adjustbox}{width=\textwidth}
\begin{tabular}{lllrrrrrrrr}
\multicolumn{2}{c}{\textbf{Dataset}} & \multicolumn{1}{c}{\begin{tabular}[c]{@{}c@{}}\textbf{Propagation} \\ \textbf{Technique}\end{tabular}} & \multicolumn{1}{c}{\begin{tabular}[c]{@{}c@{}}$\mid S \mid$\end{tabular}} & \multicolumn{1}{c}{\begin{tabular}[c]{@{}c@{}}\textbf{Average}\\ $\mid L_i \cup L_c \mid$\end{tabular}} & \multicolumn{1}{c}{\begin{tabular}[c]{@{}c@{}}\textbf{Average}\\ \textbf{Propagation}\\ \textbf{Accuracy}\end{tabular}} & \multicolumn{1}{c}{\begin{tabular}[c]{@{}c@{}}\textbf{Average}\\ $\mid S\cup L_i \cup L_c \mid$\end{tabular}} & \multicolumn{1}{c}{\begin{tabular}[c]{@{}c@{}}\textbf{Average} $\kappa$ \\ \textbf{(SVM)}\end{tabular}} & \multicolumn{1}{c}{\begin{tabular}[c]{@{}c@{}}\textbf{Average} $\kappa$ \\ \textbf{(OPF)}\end{tabular}}  \\
\hline
\multicolumn{2}{l}{\multirow{3}{*}{MNIST}}  &    OPF-Semi (2D)  &    175    &    3325   &   0.796592    &    3500   &    0.780197 $\pm$ 0.008   &    0.751794 $\pm$ 0.010     \\ 
\multicolumn{2}{l}{}    &    ILP    &    175    &    1864   &    \textbf{0.974718}  &    2039   &    0.844264 $\pm$ 0.027   &    0.776241 $\pm$ 0.036   \\ 
\multicolumn{2}{l}{}    &    SALP   &   175     &    2872   &    0.947192   &    3047   &    \textbf{0.885855 $\pm$ 0.030}  &    \textbf{0.839161 $\pm$ 0.018}    \\
\hline
\multicolumn{2}{l}{\multirow{3}{*}{H.Eggs}}     &    OPF-Semi (2D)  &    61     &    1175   &    0.982993   &    1236   &    0.983621 $\pm$ 0.011     &    0.982146 $\pm$ 0.008   \\ 
\multicolumn{2}{l}{}    &    ILP    &    61     &   1171    &    \textbf{0.996014}  &    1232   &   \textbf{0.986624 $\pm$ 0.009}   &    \textbf{0.987364 $\pm$ 0.003}   \\ 
\multicolumn{2}{l}{}    &   SALP    &    61     &    1175   &    0.992347   &    1236   &    \textbf{0.989582 $\pm$ 0.005}  &    \textbf{0.983639 $\pm$ 0.007}    \\
\hline
\multicolumn{2}{l}{\multirow{3}{*}{P.cysts}}    &    OPF-Semi (2D)  &    134    &    2562   &    0.838017   &    2696   &    0.801953 $\pm$ 0.021     &    0.786383 $\pm$ 0.022   \\    
\multicolumn{2}{l}{}    &    ILP    &    134    &    1999   &    0.947177   &    2133   &    0.851948 $\pm$ 0.006   &    0.841023 $\pm$ 0.002   \\  
\multicolumn{2}{l}{}    &   SALP    &    134    &    2309   &    \textbf{0.951119}  &    2443   &    \textbf{0.877566 $\pm$ 0.011}  &    \textbf{0.850232 $\pm$ 0.015}   \\
\hline
\multicolumn{2}{l}{\multirow{3}{*}{H.Larvae (I)}}   &    OPF-Semi (2D)  &    122    &    2337   &    0.924405   &    2459   &    0.569164 $\pm$ 0.070   &    0.556642 $\pm$ 0.059   \\
\multicolumn{2}{l}{}    &    ILP    &    122    &    2080   &    0.981273   &    2202   &    0.727843 $\pm$ 0.013   &    0.723049 $\pm$ 0.016   \\
\multicolumn{2}{l}{}    &    SALP   &    122    &    2337   &    \textbf{0.986730}  &    2459   &    \textbf{0.805388 $\pm$ 0.014}  &    \textbf{0.748340 $\pm$ 0.036}   \\
\hline
\multicolumn{2}{l}{\multirow{3}{*}{H.Eggs (I)}}     &    OPF-Semi (2D)  &    178    &    3400   &    0.729608   &    3578   &    0.611144 $\pm$ 0.071   &    0.544552 $\pm$ 0.047   \\  
\multicolumn{2}{l}{}    &    ILP    &    178    &    1547   &   0.914358    &    1725   &    0.683544 $\pm$ 0.033   &    0.593104 $\pm$ 0.034   \\ 
\multicolumn{2}{l}{}    &    SALP   &    178    &    3059   &    \textbf{0.959611}  &    3237   &    \textbf{0.866121 $\pm$ 0.043}  &    \textbf{0.725803 $\pm$ 0.025}   \\
\hline
\multicolumn{2}{l}{\multirow{3}{*}{P.cysts (I)}}    &    OPF-Semi (2D)  &    334    &    6363   &    0.605427   &    6697   &    0.429731 $\pm$ 0.013   &    0.396645 $\pm$ 0.009   \\ 
\multicolumn{2}{l}{}    &    ILP    &    334    &    1787   &    0.826867   &    2121   &    0.589643 $\pm$ 0.036    &   0.472148 $\pm$ 0.008   \\ 
\multicolumn{2}{l}{}    &    SALP   &    334    &    3948   &    \textbf{0.864390}  &    4282   &    \textbf{0.648831 $\pm$ 0.043}  &    \textbf{0.543963 $\pm$ 0.016}   \\ 
\hline
\end{tabular}
\end{adjustbox}
\end{table*}

%% file: graphs/nd_2d.tex
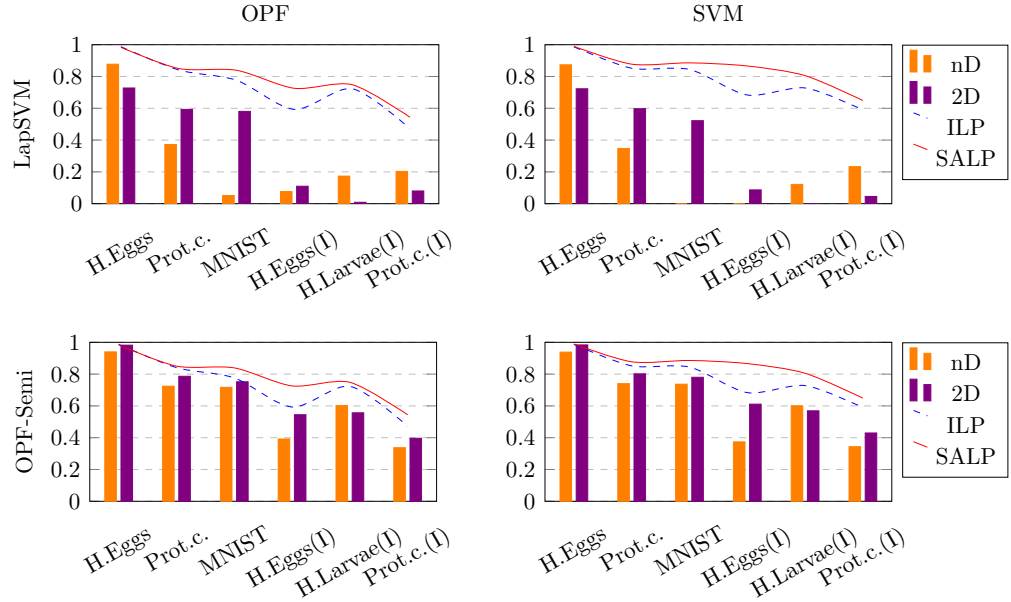
\begin{figure}[h]
\begin{center}
\begin{adjustbox}{width=\textwidth}
\begin{tabular}{cc}
\begin{tikzpicture}
    \begin{axis}[
        title={\footnotesize OPF},
        width  = 0.5*\textwidth,
        height = 4cm,
        major x tick style = transparent,
        ticklabel style = {font=\footnotesize},
        label style = {font=\footnotesize},
        ybar,
        bar width=5pt,
        ymajorgrids = true,
        ylabel = {\footnotesize LapSVM},
        symbolic x coords={H.Eggs,Prot.c.,MNIST,H.Eggs(I),H.Larvae(I),Prot.c.(I)},
        xtick = data,
        xticklabel style={rotate=30},
        scaled y ticks = false,
        ymin=0,
        ymax=1,
        grid style=dashed,
        ytick = {0, 0.2, 0.4, 0.6, 0.8, 1},
    ]
        \addplot[style={orange,fill=orange,mark=none}]
            coordinates {(H.Eggs,0.877344) (Prot.c.,0.371770) (MNIST,0.051110) (H.Eggs(I),0.076043) (H.Larvae(I),0.173416) (Prot.c.(I),0.202826)};
        \addplot[style={violet,fill=violet,mark=none}]
            coordinates {(H.Eggs,0.727673) (Prot.c.,0.592956) (MNIST,0.580445) (H.Eggs(I),0.109734) (H.Larvae(I),0.008665) (Prot.c.(I),0.079854)};
        \addplot[draw=blue, smooth, dashed]
            coordinates {(H.Eggs,0.987364)
             (Prot.c.,0.841023) (MNIST,0.776241) (H.Eggs(I),0.593104) (H.Larvae(I),0.723049) (Prot.c.(I),0.472148)};
        \addplot[draw=red, smooth]
            coordinates {(H.Eggs,0.983639)
             (Prot.c.,0.850232) (MNIST,0.839161) (H.Eggs(I),0.725803)  (H.Larvae(I),0.748340)(Prot.c.(I),0.543963)};
    \end{axis}
\end{tikzpicture}
&
\begin{tikzpicture}
    \begin{axis}[
        title={\footnotesize SVM},
        width  = 0.5*\textwidth,
        height = 4cm,
        major x tick style = transparent,
        ticklabel style = {font=\footnotesize},
        label style = {font=\footnotesize},
        ybar,
        bar width=5pt,
        ymajorgrids = true,
        ylabel = {},
        symbolic x coords={H.Eggs,Prot.c.,MNIST,H.Eggs(I),H.Larvae(I),Prot.c.(I)},
        xtick = data,
        xticklabel style={rotate=30},
        scaled y ticks = false,
        ymin=0,
        ymax=1,
        grid style=dashed,
        ytick = {0, 0.2, 0.4, 0.6, 0.8, 1},
        legend pos=outer north east
    ]
        \addplot[style={orange,fill=orange,mark=none}]
            coordinates {(H.Eggs,0.873472) (Prot.c.,0.346761) (MNIST,0.000000) (H.Eggs(I),0.000000) (H.Larvae(I),0.121253) (Prot.c.(I),0.232800)};
            \addlegendentry{\footnotesize nD}
        \addplot[style={violet,fill=violet,mark=none}]
            coordinates {(H.Eggs,0.722691) (Prot.c.,0.597541) (MNIST,0.521970) (H.Eggs(I),0.086608) (H.Larvae(I),0.000000) (Prot.c.(I),0.045538)};
            \addlegendentry{\footnotesize 2D}
        \addplot[draw=blue, smooth, dashed]
            coordinates {(H.Eggs, 0.986624)
             (Prot.c.,0.851948) (MNIST,0.844264) (H.Eggs(I),0.683544) (H.Larvae(I),0.727843) (Prot.c.(I),0.589643)};
             \addlegendentry{\footnotesize ILP}
        \addplot[draw=red, smooth]
            coordinates {(H.Eggs,0.989582)
             (Prot.c.,0.877566) (MNIST,0.885855) (H.Eggs(I),0.866121) (H.Larvae(I),0.805388)  (Prot.c.(I),0.648831)};
             \addlegendentry{\footnotesize SALP}
    \end{axis}
\end{tikzpicture} \\

\begin{tikzpicture}
    \begin{axis}[
        width  = 0.5*\textwidth,
        height = 4cm,
        major x tick style = transparent,
        ticklabel style = {font=\footnotesize},
        label style = {font=\footnotesize},
        ybar,
        bar width=5pt,
        ymajorgrids = true,
        ylabel = {\footnotesize OPF-Semi},
        symbolic x coords={H.Eggs,Prot.c.,MNIST,H.Eggs(I),H.Larvae(I),Prot.c.(I)},
        xtick = data,
        xticklabel style={rotate=30},
        scaled y ticks = false,
        ymin=0,
        ymax=1,
        grid style=dashed,
        ytick = {0, 0.2, 0.4, 0.6, 0.8, 1},
    ]
        \addplot[style={orange,fill=orange,mark=none}]
            coordinates {(H.Eggs,0.939834) (Prot.c.,0.724182) (MNIST,0.716913) (H.Eggs(I),0.391894) (H.Larvae(I),0.602001) (Prot.c.(I),0.337168)};
        \addplot[style={violet,fill=violet,mark=none}]
            coordinates {(H.Eggs,0.982146) (Prot.c.,0.786383) (MNIST,0.751794) (H.Eggs(I),0.544552) (H.Larvae(I),0.556642) (Prot.c.(I),0.396645)};
        \addplot[draw=blue, smooth, dashed]
            coordinates {(H.Eggs,0.987364)
             (Prot.c.,0.841023) (MNIST,0.776241) (H.Eggs(I),0.593104) (H.Larvae(I),0.723049) (Prot.c.(I),0.472148)};
        \addplot[draw=red, smooth]
            coordinates {(H.Eggs,0.983639)
             (Prot.c.,0.850232) (MNIST,0.839161) (H.Eggs(I),0.725803) (H.Larvae(I),0.748340) (Prot.c.(I),0.543963)};
    \end{axis}
\end{tikzpicture} &

\begin{tikzpicture}
    \begin{axis}[
        width  = 0.5*\textwidth,
        height = 4cm,
        major x tick style = transparent,
        ticklabel style = {font=\footnotesize},
        label style = {font=\footnotesize},
        ybar,
        bar width=5pt,
        ymajorgrids = true,
        ylabel = {},
        symbolic x coords={H.Eggs,Prot.c.,MNIST,H.Eggs(I),H.Larvae(I),Prot.c.(I)},
        xtick = data,
        xticklabel style={rotate=30},
        scaled y ticks = false,
        ymin=0,
        ymax=1,
        grid style=dashed,
        ytick = {0, 0.2, 0.4, 0.6, 0.8, 1},
        legend pos=outer north east
    ]
        \addplot[style={orange,fill=orange,mark=none}]
            coordinates {(H.Eggs,0.938317) (Prot.c.,0.740287) (MNIST,0.736685) (H.Eggs(I),0.373763) (H.Larvae(I),0.601003) (Prot.c.(I),0.343356)};
            \addlegendentry{\footnotesize nD}
        \addplot[style={violet,fill=violet,mark=none}]
            coordinates {(H.Eggs,0.983621) (Prot.c.,0.801953) (MNIST,0.780197) (H.Eggs(I),0.611144) (H.Larvae(I),0.569164) (Prot.c.(I),0.429731)};
            \addlegendentry{\footnotesize 2D}
        \addplot[draw=blue, smooth, dashed]
            coordinates {(H.Eggs, 0.986624)
             (Prot.c.,0.851948) (MNIST,0.844264) (H.Eggs(I),0.683544) (H.Larvae(I),0.727843) (Prot.c.(I),0.589643)};
             \addlegendentry{\footnotesize ILP}
        \addplot[draw=red, smooth]
            coordinates {(H.Eggs,0.989582)
             (Prot.c.,0.877566) (MNIST,0.885855) (H.Eggs(I),0.866121) (H.Larvae(I),0.805388) (Prot.c.(I),0.648831)};
             \addlegendentry{\footnotesize SALP}
    \end{axis}
\end{tikzpicture}\\

\end{tabular}
\end{adjustbox}
\end{center}
\caption{$\kappa$ values for studied datasets, for OPF and SVM classifiers (columns) and LapSVM and OPF-Semi automatic label propagation methods (rows). Curves show $\kappa$ values for ILP and SALP for comparison purposes. Datasets are sorted from easiest to hardest to classify (left to right) based on SALP results.
}
    \label{f.2d_nd_feat}
\end{figure}

%% file: graphs/user_effort.tex
\begin{figure}[h]
    \centering
\begin{adjustbox}{width=0.5\textwidth}
\begin{tabular}{c}
\begin{tikzpicture}
    \begin{axis}[
        title={\footnotesize Percentage of labeled samples as function of different datasets},
        width  = 0.6*\textwidth,
        height = 5cm,
        major x tick style = transparent,
        ticklabel style = {font=\footnotesize},
        label style = {font=\footnotesize},
        ybar,
        bar width=6pt,
        ymajorgrids = true,
        ylabel = {\footnotesize \% of labeled samples},
        symbolic x coords={H.Eggs,H.Larvae(I),Prot.c.,MNIST,H.Eggs(I),Prot.c.(I)},
        xticklabel style={rotate=30},
        scaled y ticks = false,
        ymin=0,
        ymax=1,
        grid style=dashed,
        legend pos=outer north east
    ]
        \addplot[style={blue,fill=blue,mark=none}]
            coordinates {(H.Eggs,0.99575) (H.Larvae(I),0.89003) (Prot.c.,0.78025) (MNIST,0.56060) (H.Eggs(I),0.45400) (Prot.c.(I),0.28084)};
            \addlegendentry{\footnotesize ILP}
        \addplot[style={red,fill=red,mark=none}]
            coordinates {(H.Eggs,0.130952381)  (H.Larvae(I),0.223791185) (Prot.c.,0.259953162) (MNIST,0.35518797) (H.Eggs(I),0.610588235) (Prot.c.(I),0.272355807)};
            \addlegendentry{\footnotesize SALP}
    \end{axis}
\end{tikzpicture}
\\ (a) \\
\begin{tikzpicture}
    \begin{axis}[
        title={\footnotesize Number of samples as function of different datasets},
        width  = 0.6*\textwidth,
        height = 5cm,
        major x tick style = transparent,
        ticklabel style = {font=\footnotesize},
        label style = {font=\footnotesize},
        ybar,
        bar width=6pt,
        ymajorgrids = true,
        ylabel = {\footnotesize $\#$ of samples},
        symbolic x coords={H.Eggs,H.Larvae(I),Prot.c.,MNIST,H.Eggs(I),Prot.c.(I)},
        xticklabel style={rotate=30},
        scaled y ticks = false,
        ymin=0,
        ymax=7000,
        grid style=dashed,
        legend pos=outer north east,
        ytick = {0, 1000, 2000, 3000, 4000, 5000, 6000, 7000},
    ]
        \addplot[style={orange,fill=orange,mark=none}]
            coordinates {(H.Eggs,1236) (H.Larvae(I),2459) (Prot.c.,2636) (MNIST,3500) (H.Eggs(I),3578) (Prot.c.(I),6697)};
    \end{axis}
\end{tikzpicture} \\
(b)
\end{tabular}
\end{adjustbox}
    \caption{(a) the percentage of labeled samples in $U$ \emph{vs} ILP and SALP label propagation methods and (b) the number of samples for the six studied datasets.}
    \label{f.user_effort}
\end{figure}
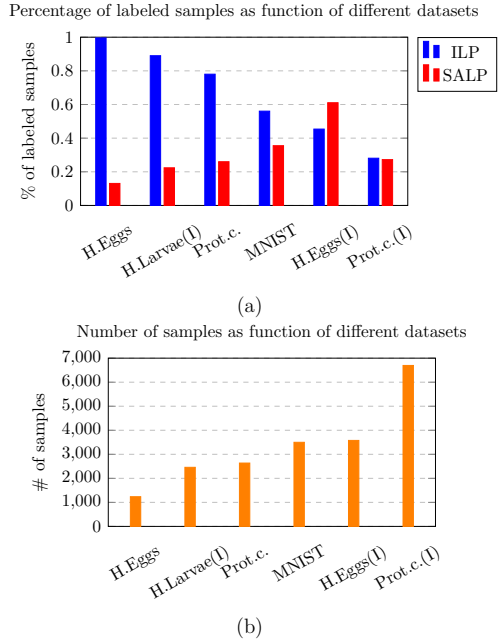

%% file: graphs/effectiveness.tex
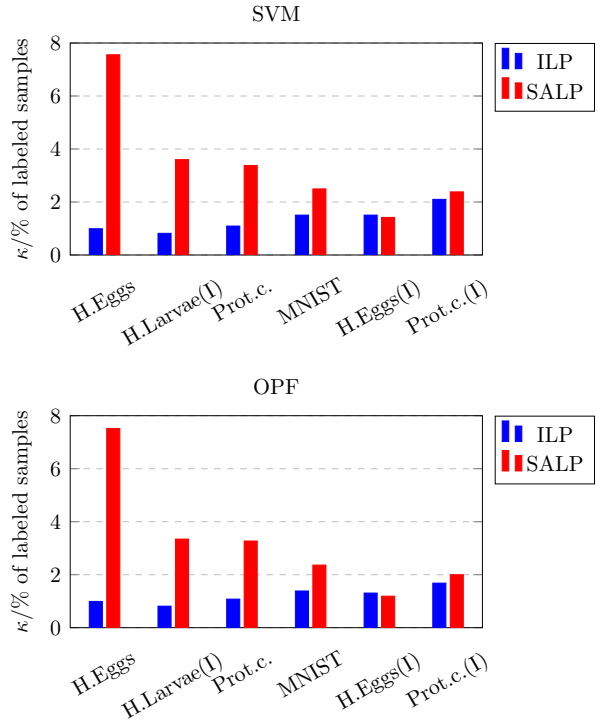
\begin{figure}[h]
    \centering
\begin{adjustbox}{width=0.6\textwidth}
\begin{tabular}{c}

\begin{tikzpicture}
    \begin{axis}[
        title={\footnotesize SVM},
        width  = 0.6*\textwidth,
        height = 5cm,
        major x tick style = transparent,
        ticklabel style = {font=\footnotesize},
        label style = {font=\footnotesize},
        ybar,
        bar width=6pt,
        ymajorgrids = true,
        ylabel = {\footnotesize$\kappa / \%$ of labeled samples},
        symbolic x coords={H.Eggs,H.Larvae(I),Prot.c.,MNIST,H.Eggs(I),Prot.c.(I)},
        xticklabel style={rotate=30},
        scaled y ticks = false,
        ymin=0,
        ymax=8,
        grid style=dashed,
        legend pos=outer north east
    ]
        \addplot[style={blue,fill=blue,mark=none}]
            coordinates {(H.Eggs,0.99084) (H.Larvae(I),0.81777) (Prot.c.,1.09189) (MNIST,1.50600) (H.Eggs(I),1.50560) (Prot.c.(I),2.09957)};
            \addlegendentry{\footnotesize ILP}
        \addplot[style={red,fill=red,mark=none}]
            coordinates {(H.Eggs,7.55681) (H.Larvae(I),3.59884) (Prot.c.,3.37586) (MNIST,2.49405) (H.Eggs(I),1.41850) (Prot.c.(I),2.38229)};
            \addlegendentry{\footnotesize SALP}
    \end{axis}
\end{tikzpicture}
\\

\begin{tikzpicture}
    \begin{axis}[
        title={\footnotesize OPF},
        width  = 0.6*\textwidth,
        height = 5cm,
        major x tick style = transparent,
        ticklabel style = {font=\footnotesize},
        label style = {font=\footnotesize},
        ybar,
        bar width=6pt,
        ymajorgrids = true,
        ylabel = {\footnotesize $\kappa / \%$ of labeled samples},
        symbolic x coords={H.Eggs,H.Larvae(I),Prot.c.,MNIST,H.Eggs(I),Prot.c.(I)},
        xticklabel style={rotate=30},
        scaled y ticks = false,
        ymin=0,
        ymax=8,
        grid style=dashed,
        legend pos=outer north east
    ]
        \addplot[style={blue,fill=blue,mark=none}]
            coordinates {(H.Eggs,0.9916) (H.Larvae(I),0.81239) (Prot.c.,1.07789) (MNIST,1.38466) (H.Eggs(I),1.30640) (Prot.c.(I),1.68120)};
            \addlegendentry{\footnotesize ILP}
        \addplot[style={red,fill=red,mark=none}]
            coordinates {(H.Eggs,7.51425)  (H.Larvae(I),3.34392) (Prot.c.,3.27071) (MNIST,2.36258) (H.Eggs(I),1.18869) (Prot.c.(I),1.99725)};
            \addlegendentry{\footnotesize SALP}
    \end{axis}
\end{tikzpicture}\\
\end{tabular}
\end{adjustbox}
    \caption{Normalized gain, i.e., $\kappa$ divided by the percentage of manually labeled samples with ILP and SALP for SVM and OPF classifiers for the six studied datasets (sorted left to right on size).}
    \label{f.effectiveness}
\end{figure}